\documentclass[10pt,journal,compsoc]{IEEEtran}

\ifCLASSOPTIONcompsoc
  \usepackage[nocompress]{cite}
\else
  \usepackage{cite}
\fi

\usepackage{graphicx}
\usepackage{ragged2e}
\usepackage{multirow}
\usepackage{amssymb}
\usepackage{amsmath}
\usepackage{booktabs}
\usepackage[switch]{lineno}
\usepackage[dvipsnames,svgnames,table]{xcolor}
\usepackage{amsmath,amssymb}
\usepackage[ruled,linesnumbered]{algorithm2e}

\SetArgSty{textnormal}
\SetKwInOut{Input}{Input}
\SetKwInOut{Output}{Output}
\DontPrintSemicolon


\usepackage{color}
\usepackage{ragged2e}
\usepackage{hyperref}
\usepackage{cite}
\usepackage{array}
\usepackage{colortbl}
\usepackage{alphalph}

\usepackage{subcaption}
\usepackage[export]{adjustbox}
\usepackage{wrapfig}

\newcommand{\w}{{\mathbf{w}}}
\newcommand{\m}{{\mathbf{m}}}
\newcommand{\z}{{\mathbf{z}}}

\ifCLASSINFOpdf
\else
\fi

\renewcommand{\justify}{\leftskip=0pt \rightskip=0pt plus 0cm}

\begin{document}

\title{Lottery Jackpots Exist in Pre-trained Models}

\author{Yuxin Zhang,
        Mingbao Lin,
        Yunshan Zhong,
        Fei Chao,~\IEEEmembership{Member,~IEEE},
        Rongrong Ji,~\IEEEmembership{Senior Member,~IEEE}
\IEEEcompsocitemizethanks{\IEEEcompsocthanksitem Y. Zhang, Y. Zhong, F. Chao, and R. Ji (Corresponding Author) are with the Key Laboratory of Multimedia Trusted Perception and Efficient Computing, Ministry of Education of China, Xiamen University, Xiamen 361005, China, and also with School of Informatics, Xiamen University, Xiamen 361005, China (e-mail: rrji@xmu.edu.cn).
\IEEEcompsocthanksitem M. Lin is with Youtu Laboratory, Tencent, Shanghai 200233, China.
\IEEEcompsocthanksitem R. Ji is also with Institute of Artificial Intelligence, Xiamen University, Xiamen 361005, China. 
}
\thanks{Manuscript received April 19, 2005; revised August 26, 2015.}}

\markboth{IEEE TRANSACTIONS ON PATTERN ANALYSIS AND MACHINE INTELLIGENCE}%
{Shell \MakeLowercase{\textit{et al.}}: Bare Demo of IEEEtran.cls for IEEE Journals}

\IEEEtitleabstractindextext{%
\begin{abstract}
\justify{Network pruning is an effective approach to reduce network complexity with acceptable performance compromise. Existing studies achieve the sparsity of neural networks via time-consuming weight training} or complex searching on networks with expanded width, which greatly limits the applications of network pruning. In this paper, we show that high-performing and sparse sub-networks without the involvement of weight training, termed ``lottery jackpots'', exist in pre-trained models with unexpanded width. 
Our presented lottery jackpots are traceable through empirical and theoretical outcomes.
For example, we obtain a lottery jackpot that has only 10\% parameters and still reaches the performance of the original dense VGGNet-19 without any modifications on the pre-trained weights on CIFAR-10.
Furthermore, we improve the efficiency for searching lottery jackpots from two perspectives.
Firstly, we observe that the sparse masks derived from many existing pruning criteria have a high overlap with the searched mask of our lottery jackpot, among which, the magnitude-based pruning results in the most similar mask with ours. 
In compliance with this insight, we initialize our sparse mask using the magnitude-based pruning, resulting in at least 3$\times$ cost reduction on the lottery jackpot searching while achieving comparable or even better performance.
Secondly, we conduct an in-depth analysis of the searching process for lottery jackpots.
Our theoretical result suggests that the decrease in training loss during weight searching can be disturbed by the dependency between weights in modern networks.
To mitigate this, we propose a novel short restriction method to restrict change of masks that may have potential negative impacts on the training loss, which leads to a faster convergence and reduced oscillation for searching lottery jackpots.
Consequently, our searched lottery jackpot removes 90\% weights in ResNet-50, while it easily obtains more than 70\% top-1 accuracy using only 5 searching epochs on ImageNet. Our code is available at \url{https://github.com/zyxxmu/lottery-jackpots}.
\end{abstract}

}

\maketitle


\IEEEpeerreviewmaketitle

\section{Introduction}
\label{sec:intro}
\IEEEPARstart{E}{ver-increasing} model complexity has greatly limited the real-world applications of deep neural networks (DNNs) on edge devices. Various methods have been proposed to mitigate this obstacle by the deep learning community. Generally, existing research can be divided into network pruning~\cite{han2015learning,he2019filter}, parameter quantization~\cite{hubara2016binarized, zhong2022fine,zhong2022intraq}, low-rank decomposition~\cite{peng2018extreme, hayashi2019exploring} and knowledge distillation~\cite{romero2014fitnets,hinton2015distilling}. Among these techniques, network pruning has been known as one of the leading approaches with notable reductions on the network complexity and acceptable performance degradation~\cite{ashbyexploiting,elsen2020fast,park2016holistic}.

\begin{figure}[!t]
\begin{center}
\includegraphics[height=0.65\linewidth]{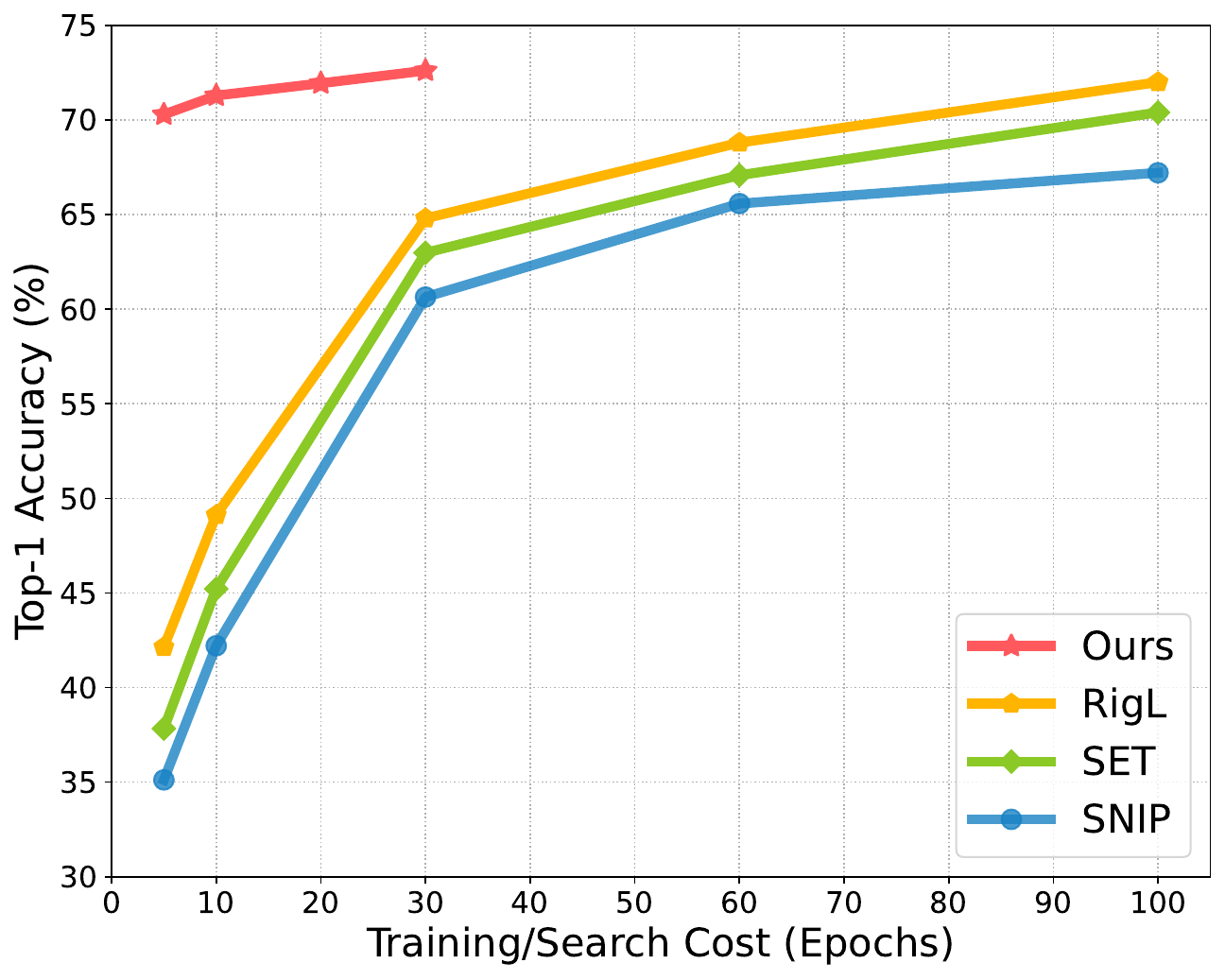}
\end{center}
\caption{\label{comparison}
Training/Search cost \emph{v.s.} top-1 accuracy of ResNet-50~\cite{he2016deep} with a sparse rate of $90\%$ on ImageNet~\cite{deng2009imagenet}. Search epoch differs from training epoch in that it only trains the mask for indicating the removal or preserve of weights, without modifying the weight value. Our method can quickly find the high-performing sparse networks (lottery jackpots) without modifications on the trained weights, while existing methods such as SET~\cite{mocanu2018scalable}, SNIP~\cite{lee2018snip} and RigL~\cite{evci2020rigging} conduct pruning via a time-consuming weight training process to recover the performance. Each dot in the figure indicates a complete training with cosine annealing. 
}
\end{figure}

Given a large-scale neural network, network pruning removes a portion of network connections to obtain a sparse sub-network. Extensive pruning algorithms have been proposed over the past few years~\cite{molchanov2017variational,chang2020provable,joo2021linearly}. Typical approaches devise various importance criteria to prune weights on the basis of pre-trained models, which is reasonable since the pre-trained models are mostly visible on the Internet, or available from the client~\cite{lecun1989optimal,han2015learning,he2020learning}.
Other studies conduct pruning from scratch by imposing a sparsity regularization upon the network loss, or directly pruning from randomly initialized weights~\cite{mostafa2019parameter,evci2020rigging,wang2020pruning}. Although progress has been made to reduce the size of network parameters with little degradation in accuracy, existing methods still require a time-consuming weight training process to recover the performance of pruned models as shown in Fig.\,\ref{comparison}. For instance, when pruning the well-known ResNet-50~\cite{he2016deep} on ImageNet~\cite{deng2009imagenet}, most methods require over $100$ epochs to train the pruned model~\cite{mostafa2019parameter,kusupati2020soft,evci2020rigging}. Thus, the compressed models come at the cost of expensive weight training, which greatly restricts practical applications of existing researches.

%
The lottery ticket hypothesis~\cite{frankle2018lottery} reveals that a randomly initialized network contains lottery ticket sub-networks that can reach good performance after appropriate weight training. In light of this, more recent studies further discover that these lottery tickets emerge even without the necessity of weight training~\cite{ramanujan2020s,zhou2019deconstructing,orseau2020logarithmic}. However, a complex searching algorithm has to be conducted upon randomly initialized weights, the cost of which is even higher than training the weights since the width of the original network is usually exponentially expanded to ensure the finding of high-performing lottery tickets.
Moreover, the performance of such sub-networks still falls far behind existing weight-training pruning methods~\cite{mostafa2019parameter,kusupati2020soft,evci2020rigging}.

In this paper, we innovatively reveal that high-performing sub-networks can be located in pre-trained models without the involvement of weight training from empirical and theoretical perspectives.
We term these sub-networks as lottery jackpots in this paper. First, different from existing works~\cite{ramanujan2020s,zhou2019deconstructing,orseau2020logarithmic} which search for the lottery tickets in a randomly initialized network with width expansion, our lottery jackpots are built on top of an unexpanded pre-trained model. For example, a lottery jackpot can be found in the pre-trained VGGNet-19~\cite{simonyan2015very} on CIFAR-10~\cite{krizhevsky2009learning}, which has only 10\% parameters of the original network while still reaching the performance of the full model without any modifications on the trained weights.
Nevertheless, off-the-shelf edge-popup algorithm~\cite{ramanujan2020s} that we leverage to find lottery jackpots is time-consuming. It takes almost the same computation cost compared with existing weight training methods~\cite{lee2018snip, wang2020picking}, which heavily barricades the application value of lottery jackpots.

To alleviate the above problem, we further propose to improve the searching efficiency of lottery jackpots from two perspectives, unfolded as mask initialization and mask search.
Toward the first goal, we look into existing weight-training based pruning criteria~\cite{han2015learning,lee2018snip,wang2020picking}, and then observe a high overlap between the sparsity masks from these existing pruning criteria and the searched mask from our lottery jackpot. Among them, the magnitude-based pruning criterion results in the most similar sparsity pattern. This inspires us to initialize our mask with magnitude pruning as a warm-up for searching our lottery jackpots. As a result, at least $3\times$ reductions on the searching cost are observed when compared to the existing weight searching methods on the randomly initialized networks~\cite{ramanujan2020s,zhou2019deconstructing} or weight training methods~\cite{mostafa2019parameter, evci2020rigging, mocanu2018scalable}.

Next, we study how to boost the efficiency of the searching process for lottery jackpots. 
As the pre-trained weights are fixed during the searching phase, the only opportunity to decrease training loss falls into pruning weights that are preserved in the previous searching iterations and reviving back the same number of pruned weights, which we call weight swapping in this paper.
Then, we mathematically prove that the loss drop carried by such weight swapping in the edge-popup algorithm is bounded by a distortion error item with regard to the dependency between weights in modern DNNs, limiting the searching efficiency of lottery jackpots.
To cope with this drawback, we present an intuitive yet effective short restriction popup, which greedily prevents weight swapping that receives little or even negative influence for minimizing the training loss.
Consequently, faster convergence and less oscillation can be reached for searching lottery jackpots.
Extensive experiments have demonstrated that our proposed short restriction popup, termed as SR-popup, can efficiently locate high-performing lottery jackpots in many representative DNNs including ResNet~\cite{he2016deep}, MobileNet-V1~\cite{howard2017mobilenets},~\emph{etc}.
For instance, SR-popup successfully searches a lottery jackpot that removes $90\%$ weights of ResNet-50~\cite{he2016deep} while reaching the top-1 accuracy of $70\%$ using only $5$ searching epochs on ImageNet and achieves comparable performance with state-of-the-art methods using only $30$ searching epochs, as shown in Fig.\,\ref{comparison}.

Overall, our contributions are summarized as:
\begin{itemize}
\item We find that pre-trained models contain high-performing sub-networks without the necessity of weight training. Moreover, our searching for these sub-networks is built on the top of no expansion of the network width.

\item We discover that existing weight-training based pruning criteria often generate a similar sparsity mask with our searched mask. Inspired by this discovery, we propose to use magnitude-based sparsity mask as a warm-up for the weight searching, leading to a more efficient searching for our lottery jackpots.

\item  With theoretical guarantee, we propose a novel short restriction popup, which adaptively preserves mask changes that earn largest expected loss drops, such that the searching instability of a previous method edge-popup is effectively relieved.

\item Extensive experiments demonstrate the effectiveness and efficiency of our proposed approach for network pruning. The high-performing sub-networks can be expeditiously found in pre-trained models with at least $3\times$ reductions on the computation cost in comparison with the existing methods.
\end{itemize}

\section{Related Work}
\label{relat}
\textbf{Neural Network Pruning.}    
Pruning neural networks has been demonstrated to be an effective approach for compressing large models in the past decades~\cite{lecun1989optimal,hassibi1992second,thimm1995evaluating}. Earlier techniques usually implement pruning by designing various importance criteria upon pre-trained models~\cite{han2015learning,molchanov2017variational,lecun1989optimal}. For instance, Han~\emph{et al.}~\cite{han2015learning} considered the magnitude of weights and Molchanov~\emph{et al.}~\cite{molchanov2016pruning} treated Taylor expansion that approximates changes in the loss function as the pruning principle.
Recently, a plurality of studies has questioned the role of pre-trained models by devising well-performing pruning algorithms without the dependency on the pre-trained models. For example, Sparse Evolutionary Training (SET)~\cite{mocanu2018scalable} cultivates sparse weights throughout training in a prune-redistribute-regrowth manner while training the models from scratch. Kusupati~\emph{et al.}~\cite{kusupati2020soft} designed Soft Threshold Reparameterization (STR) to learn non-uniform sparsity budget across layers.
Evci~\emph{et al.}~\cite{evci2020rigging} further proposed RigL, which uses weight magnitude and gradient information to improve the sparse networks optimization.
Other works directly prune a randomly initialized model by considering the connection sensitivity as an importance measure~\cite{lee2018snip} or maximizing the gradient of the pruned sub-network~\cite{wang2020picking}.

Network pruning can also be viewed as finding binary masks that indicate the removal of preserve of weights. 
For instance, Guo~\emph{et al.}~\cite{guo2016dynamic} proposed to iteratively update the binary masks and train the network parameters to avoid incorrect pruning.
Such binary mask can also be learned during training with additional gate variables~\cite{srinivas2017training} or auxiliary parameters~\cite{xiao2019autoprune}.
Savarese~\emph{et al.}~\cite{savarese2020winning} further proposed Continuous Sparsification to learn the binary masks using a differentiable approximation to $\ell_0$-regularization penalty for the parameter count of sparse networks.

There are also multiple studies that remove the entire neurons or convolution filters to achieve a high-level acceleration through parallelization~\cite{li2016pruning,ding2019centripetal,liu2019metapruning, lin2020hrank,ruan2021dpfps,ding2021resrep,guo2020dmcp,li2020eagleeye}. However, models compressed by these methods often suffer severe performance degradation, \emph{i.e.}, classification accuracy drops at a high pruning rate, and thus the complexity reduction is often very limited.
In addition, some recent studies~\cite{yu2017scalpel,mao2017exploring} presented hardware-friendly weight pruning methods to enable practical complexity reduction on off-the-shelf platforms.
In this paper, we focus on weight pruning that removes individual weights to achieve a high sparsity level.
\textbf{Lottery Ticket Hypothesis.} The lottery ticket hypothesis was originally proposed in~\cite{frankle2018lottery} which reveals the existence of sub-networks (\emph{a.k.a.},``winning tickets'') in a randomly-initialized neural network that can match the test accuracy of the original network when trained in isolation. Zhou~\emph{et al}.~\cite{zhou2019deconstructing} further proved the potential existence of lottery tickets that can achieve good performance without training the weights. Inspired by this progress, Ramanujan~\emph{et al.}~\cite{ramanujan2020s} designed an edge-popup algorithm to search for these sub-networks within the randomly initialized weights. They found a sub-network of the Wide ResNet-50~\cite{zagoruyko2016wide} that is lightly smaller than, but matches the performance of ResNet-34~\cite{he2016deep} on ImageNet. Later, such sub-networks are further confirmed by ~\cite{orseau2020logarithmic} and ~\cite{ye2020greedy}.

Unfortunately, finding the sub-networks without weight training strongly relies on the original networks to be sufficiently over-parameterized. To this end, the width of randomly initialized networks is usually exponentially expanded, which however increases the searching complexity~\cite{orseau2020logarithmic,chang2020provable}. As a result, the cost of searching these sub-networks is even higher than weight training. Moreover, the searched sub-networks are still redundant with unsatisfying accuracy, far away from the purpose of network pruning~\cite{ramanujan2020s}. Thus, it is of great urgency to find out the high-performing lottery tickets with a small searching complexity.

\section{Methodology}
\label{sec:metho}
\subsection{Preliminary}\label{preliminary}
Let the weight vector of a full network be $\mathbf{w} \in \mathbb{R}^{k}$ where $k$ is the weight size. Technically, network pruning can be viewed as applying a mask $\mathbf{m} \in \{0,1\}^k$ on $\mathbf{w}$ to indicate whether to preserve or remove some of the weights.
Given a desired global sparsity $p$, the conventional objective of network pruning can be formulated as:
\begin{equation}\label{eq1}
\begin{split}
    \min_{\mathbf{w}, \mathbf{m}} \; \mathcal{L} (\mathbf{m} \odot \mathbf{w} \; ; \mathcal{D}),  \;\;\emph{s.t.} \;\; 1 - \frac{\left\| \mathbf{m} \right\|_0}{k} \geq p,
\end {split}
\end{equation}
where $\mathcal{D}$ is the observed dataset, $\mathcal{L}$ represents the loss function,  $\left\| \cdot \right\|_0$ means the standard $\ell_0$-norm, and $\odot$ denotes the element-wise multiplication.

As can be seen from Eq.\,(\ref{eq1}), most previous methods~\cite{mocanu2018scalable,mostafa2019parameter,evci2020rigging} pursue sparse DNNs by training the given weight vector $\mathbf{w}$ while learning the mask $\mathbf{m}$. Though training the weights increases the performance of pursued sparse networks, its heavy time-consumption becomes a severe bottleneck for practical deployments as discussed before.

Inspired by the lottery ticket hypothesis~\cite{frankle2018lottery} which indicates a randomly initialized network contains sub-networks (lottery tickets) that can reach considerable accuracy, recent studies~\cite{ramanujan2020s,zhou2019deconstructing,orseau2020logarithmic} proposed to search for these lottery tickets upon the initialized weights without the necessity of weight training. Basically, their learning objective can be given in the following:
\begin{equation}\label{ticket_search}
\begin{split}
    \min_{\mathbf{m}} \; \mathcal{L} (\mathbf{m} \odot \mathbf{w} \; ; \mathcal{D}),  \;\;\emph{s.t.} \;\; 1 - \frac{\left\| \mathbf{m} \right\|_0}{k} \geq p.
\end{split}
\end{equation}

The main difference between Eq.\,(\ref{eq1}) and Eq.\,(\ref{ticket_search}) is that the $\mathbf{w}$ is regarded as a constant vector to get rid of the dependency on weight training. Nevertheless, Eq.\,(\ref{ticket_search}) cannot break the limitation of time consumption: The high-performing ticket without weight training rarely exists in a randomly initialized network, thus the network width is usually exponentially expanded to ensure the existence of lottery tickets~\cite{zhou2019deconstructing,ramanujan2020s,orseau2020logarithmic}. This increases the searching space, and thus more searching cycles are required. As a result, the searching cost is even higher than training the weights in Eq.\,(\ref{eq1}). This paper attempts to solve such a time-consuming problem. In what follows, we first reveal that the lottery tickets exist in the pre-trained models without the need for network width expansion, and then provide a fast manner for searching these lottery tickets.

%
%
%

%
%

%
\subsection{Lottery Jackpots in Pre-trained Models\label{existence}}

We aim to verify the existence of lottery tickets without weight training in the pre-trained models, termed lottery jackpots. We concentrate on pre-trained models since these models are widely visible on the Internet, or available from the client, which should be fully utilized and also eliminate the necessity of pre-training a model from scratch. 
Built upon a pre-trained model, our target for finding lottery jackpots can be re-formulated as: 
\begin{equation}\label{jackpot_obj}
\begin{split}
    \min_{\mathbf{m}} \; \mathcal{L} (\mathbf{m} \odot \tilde{\mathbf{w}} \; ; \mathcal{D}),  \;\;\emph{s.t.} \;\; 1 - \frac{\left\| \mathbf{m} \right\|_0}{k} \geq p,
\end{split}
\end{equation}
where $\tilde{\mathbf{w}}$ represents the pre-trained weight vector, which differs our method from previous work~\cite{zhou2019deconstructing,ramanujan2020s} that search for the lottery tickets in a randomly initialized network with the requirement of expanded network width.

\begin{figure*}[h]
\centering
\begin{subfigure}[t]{0.33\textwidth}
        \centering
        \includegraphics[width=\textwidth]{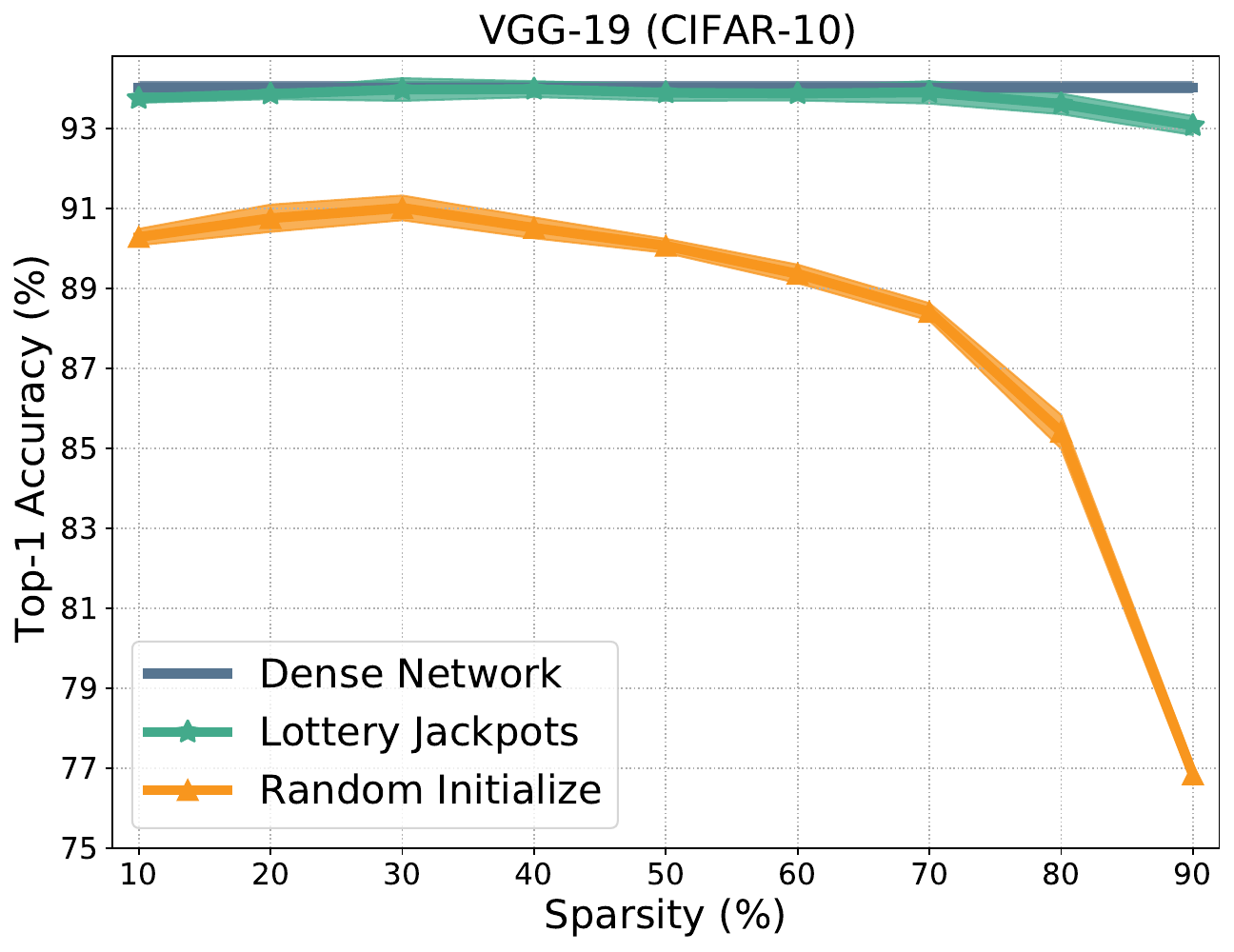}\\
\end{subfigure}
\begin{subfigure}[t]{0.33\textwidth}
        \centering
        \includegraphics[width=\textwidth]{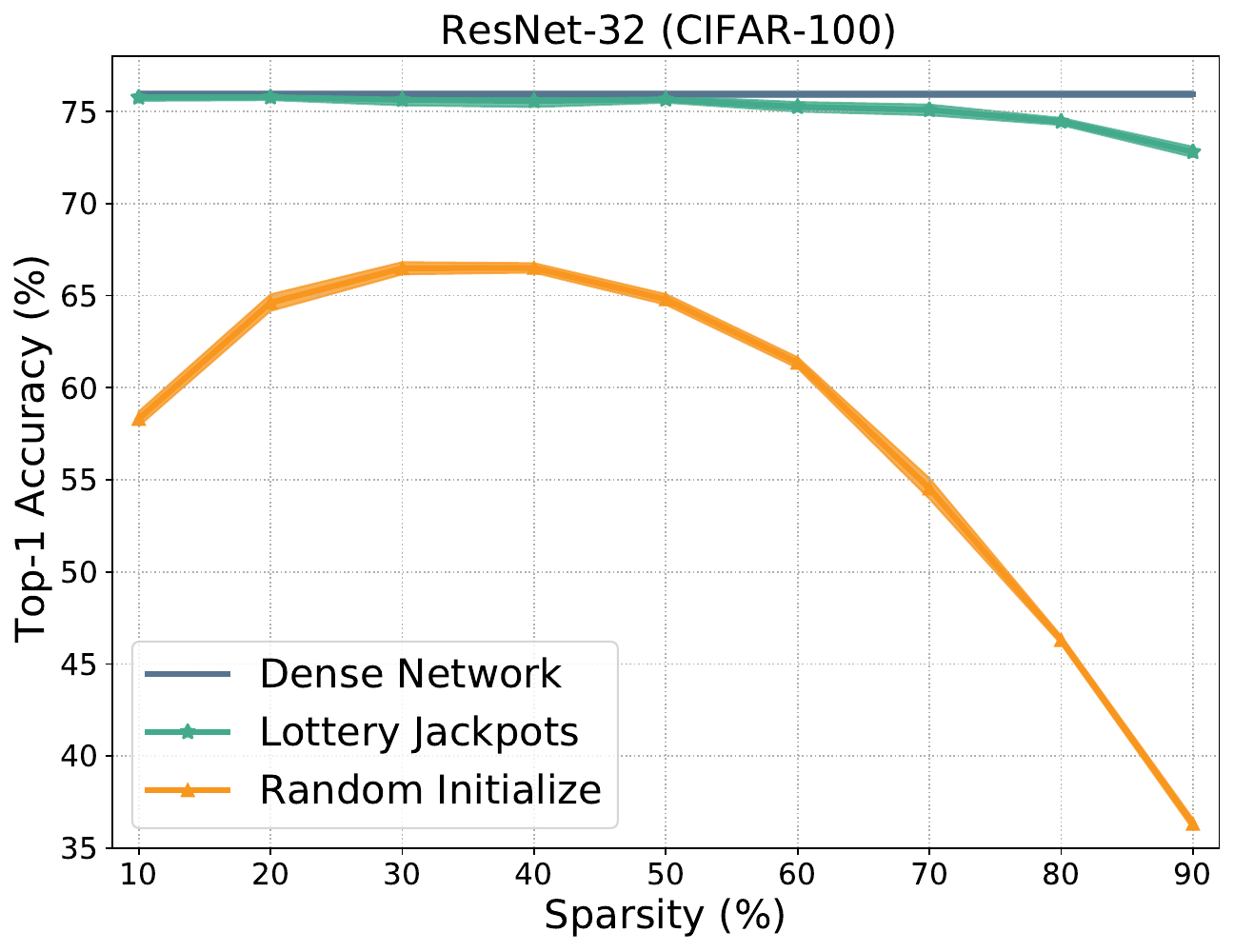}\\
\end{subfigure}
\begin{subfigure}[t]{0.33\textwidth}
        \centering
        \includegraphics[width=\textwidth]{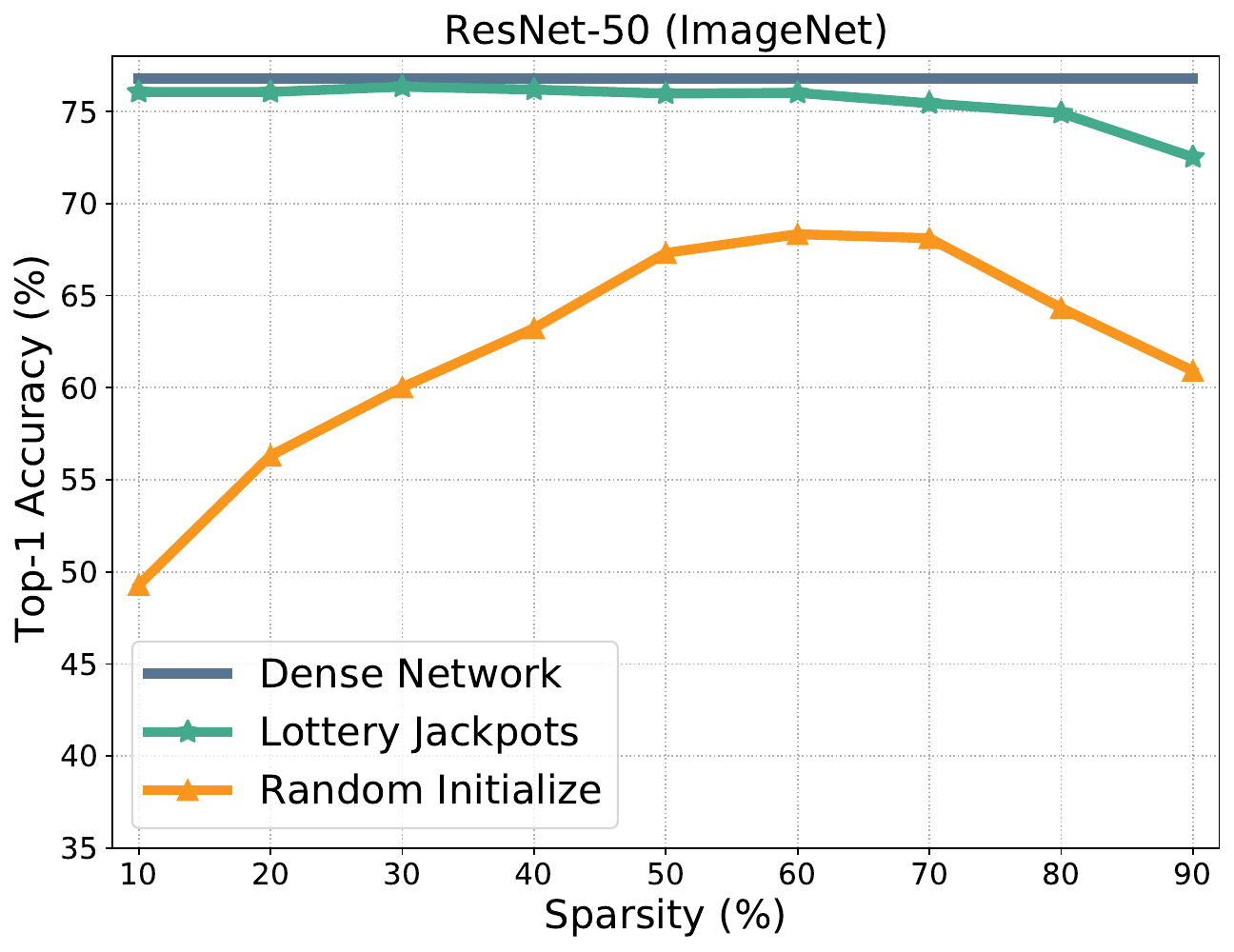}\\
\end{subfigure}
\centering
\caption{Performance of our lottery jackpots searched from pre-trained models without modifying any convergent weights and subnets searched from random initialized networks at different sparsity levels. Our lottery jackpots outperform sub-networks searched from randomly initialized networks by a large margin.}
\label{fig2}
\end{figure*}

To prove the existence of lottery jackpots in an unwidened pre-trained model, we firstly adopt the edge-popup~\cite{ramanujan2020s} to search for the lottery jackpots. Specifically, the mask $\mathbf{m} \in \{0,1\}^k$ is firstly relaxed to a randomly initialized $\mathbf{\bar{m}} \in [0,1]^k$. Denote the input of the network as $\mathbf{X}$, and then the output $\mathbf{Y}$ is obtained as
\begin{equation}
    \mathbf{Y} = \mathcal{F}( h(\mathbf{\bar{m}}) \odot \tilde{\mathbf{w}}, \mathbf{X}),
\end{equation}
where $\mathcal{F}()$ represents the neural network function, and $h(\cdot)$ is defined as:
\begin{equation}\label{h_func}
h(\mathbf{\bar{m}}^i) = \left\{ \begin{array}{ll} 
 0, \; \textrm{if $\mathbf{\bar{m}}^i$ in the top-$p\%$ smallest of $\mathbf{\bar{m}}$,}\\
 1, \; \textrm{otherwise,}
  \end{array} \right.
\end{equation}
where $i \in \{1, 2, ..., k$\}.

During the network back-propagation, the straight-through-estimator (STE)~\cite{bengio2013estimating} is used to calculate the gradient of the loss $\mathcal{L}$ \emph{w.r.t}. the relaxed mask $\mathbf{\bar{m}}$ as:
\begin{equation}\label{mask_g}
    \frac{\partial \mathcal{L}}{\partial \mathbf{\bar{m}}} = \frac{\mathcal{\partial L}}{\partial h(\mathbf{\bar{m}})}\frac{\partial h(\mathbf{\bar{m}})}{\partial \mathbf{\bar{m}}\\} \approx \frac{\mathcal{\partial L}}{\partial h(\mathbf{\bar{m}})} \cdot \textbf{1}.
\end{equation}

In this manner, we conduct mask training,~\emph{i.e.}, weight searching on various pre-trained models by optimizing $\mathbf{\bar{m}}$ with the Stochastic gradient descent (SGD) optimizer to verify the existence of lottery jackpots. As can be surprisingly observed in Fig.\,\ref{fig2}, given any sparsity level, the lottery jackpots can be found in all pre-trained networks.
For example, a lottery jackpot can be found in ResNet-32~\cite{he2016deep} pre-trained on CIFAR-10~\cite{krizhevsky2009learning}, which has only 10\% parameters while retaining over 94\% top-1 accuracy.
Lottery jackpots also outperform sub-networks searched from randomly initialized networks by a large margin, which require the original networks to be exponentially expanded for better performance~\cite{ramanujan2020s,zhou2019deconstructing,orseau2020logarithmic}.
Moreover, the found lottery jackpots can achieve comparable or even better performance than state-of-the-art methods in network pruning~\cite{lee2018snip,wang2020picking,evci2020rigging}, which will be quantitatively shown in the next section.

\textbf{Discussion.}
Some of the recent studies hold a different view that the pre-trained models are not necessary. For example, the well-known lottery ticket hypothesis~\cite{frankle2018lottery} finds that sparse networks can be trained in isolation to achieve considerable accuracy even without a pre-trained model. Liu \emph{et al.}~~\cite{liu2018rethinking} stated that inheriting weights from pre-trained models is not necessarily optimal. However, their claims are built on premise of a time-consuming weight training which helps to recover the accuracy performance. In contrast, our lottery jackpots in this paper re-justify the importance of the pre-trained models where high-performing sub-networks already exist without the necessity of weight training and network width expansion.

\textbf{Traceability of Lottery Jackpots.} 
We further delve into the principles behind lottery jackpots by initiating an error analysis in network pruning.
For simplicity, we consider one-layer full-connected network, followed by a ReLU activation function. The analysis can be easily extended to multiple-layer networks and other types of networks.
Let the input be $\z_{l-1} \in \mathbb{R}^{n_{l-1}}$ and the weights be $\w_l \in \mathbb{R}^{n_{l} \times n_{l-1}}$, where $n_{l-1}$ and $n_l$ are the neuron number of layer $l-1$ and layer $l$.
Then, the output $\z_{l} \in \mathbb{R}^{n_{l}}$ is:
\begin{equation}
    \z_l = \sigma ( \w_l \cdot \z_{l-1}),
\end{equation}
where $\sigma$ is the ReLU function.
For a compressed layer with the binary mask $\m_l$, the new output $\hat{\z}_l$ is:
\begin{equation}
    \hat{\z}_l = \sigma (( \m_l \odot \w_l ) \cdot \z_{l-1}).
\end{equation}

\textbf{Proposition 1.} Denote $\xi = ||\w_l^{i,:} - ( \m_l \odot \w_l)^{i, :}||_2$ and $z_{l-1}^{\text{Max}} = max \, |\z_{l-1}|$. For the $i$-th neuron, we have:
\begin{equation}
    0 \leq |\z_l^i - \hat{\z}_l^i| \leq \xi \sqrt{n_{l-1}} z_{l-1}^{\text{Max}}.
\end{equation}

\begin{figure}[!t]
\begin{center}
\includegraphics[height=0.65\linewidth]{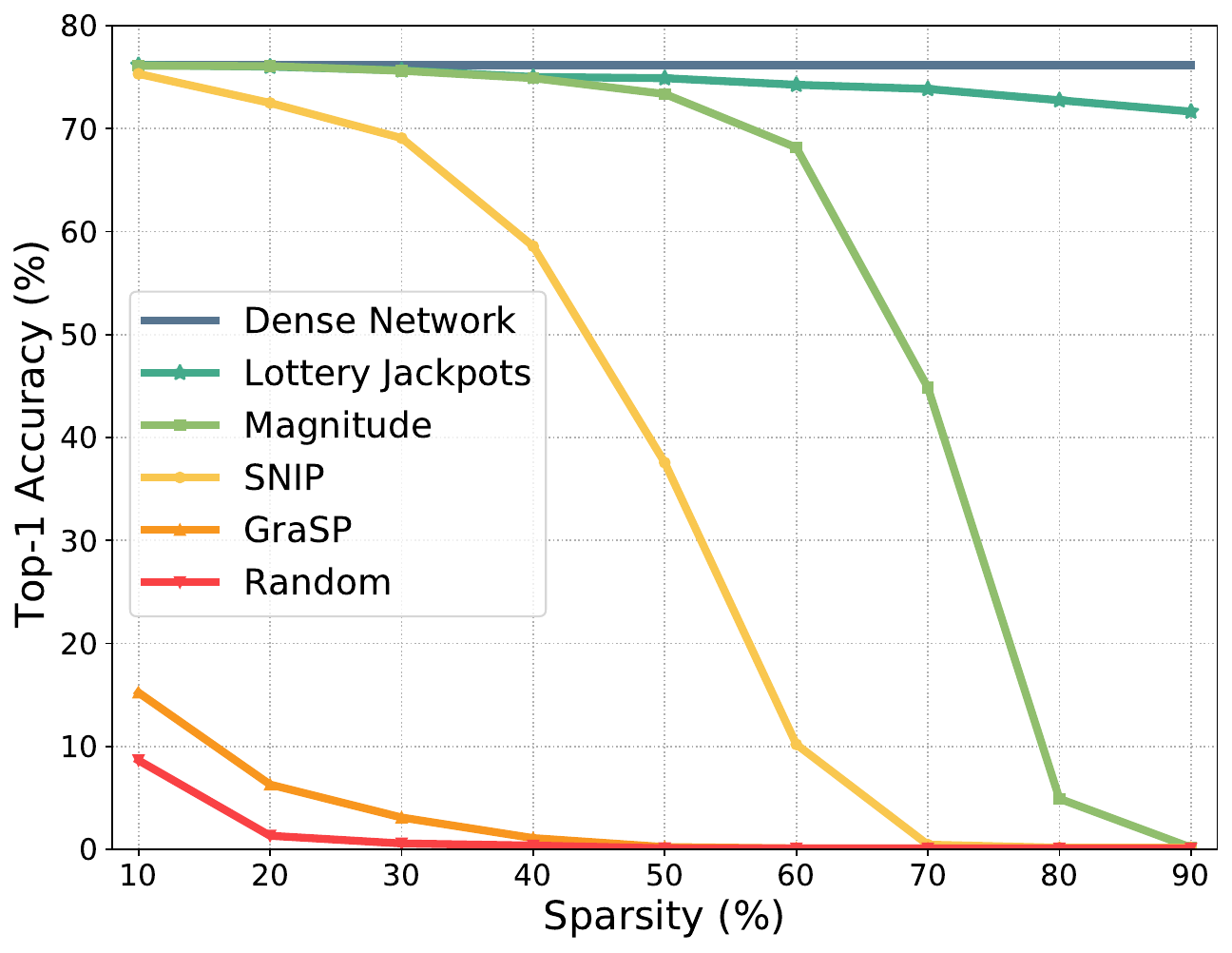}
\end{center}
\caption{\label{fig:without_retrain}
The comparison of accuracy performance between our searched lottery jackpots and existing pruning criteria without weight training. Experiments are performed using ResNet-50~\cite{he2016deep} on ImageNet~\cite{deng2009imagenet}. 
}
\end{figure}

\textbf{Proof.} $|\z_l^i - \hat{\z}_l^i|$ can be derived as:
\begin{equation}\label{eq:proposition1}
\begin{split}
\begin{aligned}
    & \quad \quad |\sigma ( \w_{l}^{i, :} \cdot \z_{l-1}) - \sigma ( ( \m_l^{i, :}  \odot \w_l^{i, :} ) \cdot  \z_{l-1}) | \quad  \quad\\
    &\overset{(a)}\leq | \w_{l}^{i, :} \cdot \z_{l-1} -  ( \m_l \odot \w_l)^{i, :} \cdot  \z_{l-1} |  \quad  \quad\\
    & \; =  | ((\textbf{1}- \m_l) \odot \w_{l})^{i, :} \cdot  \z_{l-1}  | \quad  \quad\\
    &\overset{(b)}=|Cos(((\textbf{1}- \m_l) \odot \w_{l})^{i, :},  \z_{l-1})| \quad  \quad\\
    & \quad \quad  \quad ||((\textbf{1}- \m_l) \odot \w_{l})^{i, :}||_2 ||\z_{l-1}||_2  \quad  \quad\\
    & \overset{(c)}\leq  \xi  \sqrt{n_{l-1}} \z_{l-1}^{\text{Max}} |Cos(((\textbf{1}- \m_l) \odot \w_{l})^{i,:},  \z_{l-1})|, \quad  \quad 
\end{aligned}
\end{split}
\end{equation}
where (a) follows $|\sigma(x)-\sigma(y)| \leq |x-y|$ for the ReLU functon, (b) follows $|\textbf{a} \cdot \textbf{b}| = |Cos(\textbf{a}, \textbf{b})| |\textbf{a}||_2 ||\textbf{b}||_2$ where $Cos(\textbf{a}, \textbf{b})$ returns the cosine similarity, (c) follows $||\textbf{a}||_2 \leq \sqrt{c} k$ for any $\textbf{a} \in [-k, k]^c$.
Considering $0 \leq |Cos(\textbf{a}, \textbf{b})| \leq 1$, we therefore complete the proof.
$\hfill\blacksquare$
We can observe from Eq.\,(\ref{eq:proposition1}) that the bound of output discrepancy is up to the magnitude difference $\xi$ and the cosine similarity $|Cos(((\textbf{1}- \m_l) \odot \w_{l})^{i, :},  \z_{l-1})|$. 
Taking into consideration both factors would well lower the output discrepancy.
As can be referred to Fig.~\ref{fig:without_retrain}, Magnitude-based pruning~\cite{han2015learning} fails to preserve better performance since it solely emphasizes the magnitude difference while ignoring the cosine similarity.
Nevertheless, we highlight that in situations where cosine similarity approaches zero, there even occurs no output discrepancy in network pruning, regardless of the difference in magnitude between the dense and sparse weights.

Here we give a toy example for a better understanding. Let $\w_l^{i,:} = [1, -1, 0.2, 0.3], \; \z_{l-1} = [1, 1, 1, 1]$, it is clear that magnitude-based pruning generates a binary mask of $[1, 1, 0, 0]$ at $50\%$ pruning rate to removes weights of smallest magnitude ($0.2$ and $0.3$). This produces the smallest magnitude difference $\xi = 0.3606$, cosine similarity $Cos([0,0,0.2,0.3], [1,1,1,1]) = 0.6918$, and an output derivation $|\z_l^i - \hat{\z}_l^i| = 0.5$. Nevertheless, opportunity exists in a binary mask of $[0, 0, 1, 1]$ that leads to cosine similarity $Cos([1,-1,0,0], [1,1,1,1]) = 0$. This produces an output derivation $|\z_l^i - \hat{\z}_l^i| = 0$, even with lager magnitude derivation $\xi = 1.4142$.
The above example well elucidates the rationale behind the existence of lottery jackpots in pre-trained networks.
Without any re-training on the weights, a significantly lower output discrepancy can be attained through the simultaneous consideration of both cosine similarity and magnitude difference.
Though the existence of lottery jackpots greatly highlights the values of the pre-trained models in network pruning, the weight searching still leads to a significant time consumption even though the network width is not necessary to be expanded in our lottery jackpots. For example, it takes around 100 searching epochs on ImageNet~\cite{deng2009imagenet} to successfully find the lottery jackpots. Thus, it is of great need to shorten the weight searching process to find the lottery jackpots quickly. The community might focus more on how to locate these lottery jackpots efficiently within pre-trained network models, which is also our important study in Sec.\,\ref{fast}, Sec.\,\ref{interdependence} and Sec.\,\ref{popup}.

\subsection{Mask Overlap}\label{fast}

%

%
Han~\emph{et al.}~\cite{han2015learning} observed an interesting phenomenon of ``free lunch'' that using the magnitude to prune the pre-trained AlexNet~\cite{krizhevsky2012imagenet} on ImageNet without weight training leads to no performance drops when the sparsity is less than 50\%. Despite its simplicity, the model performance degenerates sharply in a higher sparsity.  In contrast, our lottery jackpots can retain good performance at most sparsity levels as shown in Fig.\,\ref{fig2}. This inspires us to explore the potential linkage to accelerate the searching of our high-performing lottery jackpots. To this end, we first define the following ratio of sparse mask overlap to measure the similarity of two masks $\mathbf{m}_1$ and $\mathbf{m}_2$ as:
\begin{equation}
    Overlap(\mathbf{m}_1, \mathbf{m}_2) = 1 - \frac{\left\| h(\mathbf{\bar{m}}_1) - h(\mathbf{\bar{m}}_2) \right\|_1}{k}.
\end{equation}

It is easy to know that a larger overlap rate indicates two more similar masks. Then, we consider various pruning criteria in existing works based on weight-training and compare their masks with the searched mask of our lottery jackpot. We briefly revisit these typical criteria as follows:

\emph{Magnitude}~\cite{han2015learning}. This method uses weight magnitude as importance score $\mathbf{s}$ and derives a $0-1$ mask vector $\mathbf{\hat{m}}$ by removing these weights with small magnitudes.

\emph{SNIP}~\cite{lee2018snip}. The connection sensitivity to the loss is considered as weight importance score $\mathbf{s}$ and then the mask $\mathbf{\hat{m}}$ is derived by removing unimportant connections. Specifically, the importance score is defined as $\left\| \mathbf{w} \odot \mathbf{g} \right\|_1$, where $\mathbf{g}$ denotes the gradient of $\mathbf{w}$.

\emph{GraSP}~\cite{wang2020picking}. It first computes the Hessian-gradient product $\mathbf{h}$ of the $l$-th layer using the sampled training data. Then, the importance score $\mathbf{s}$ is defined as $- \mathbf{w} \odot \mathbf{h}$ to preserve the gradient flow, and the mask vector $\hat{\mathbf{m}}$ is obtained by removing low-scored weights.

\emph{Random}. We first randomly initialize the relaxed $\mathbf{\bar{m}} \in [0, 1]$, and then remove weights in compliance with these low-scored relaxed mask values.

\begin{figure}[!t]
\begin{center}
\includegraphics[height=0.65\linewidth]{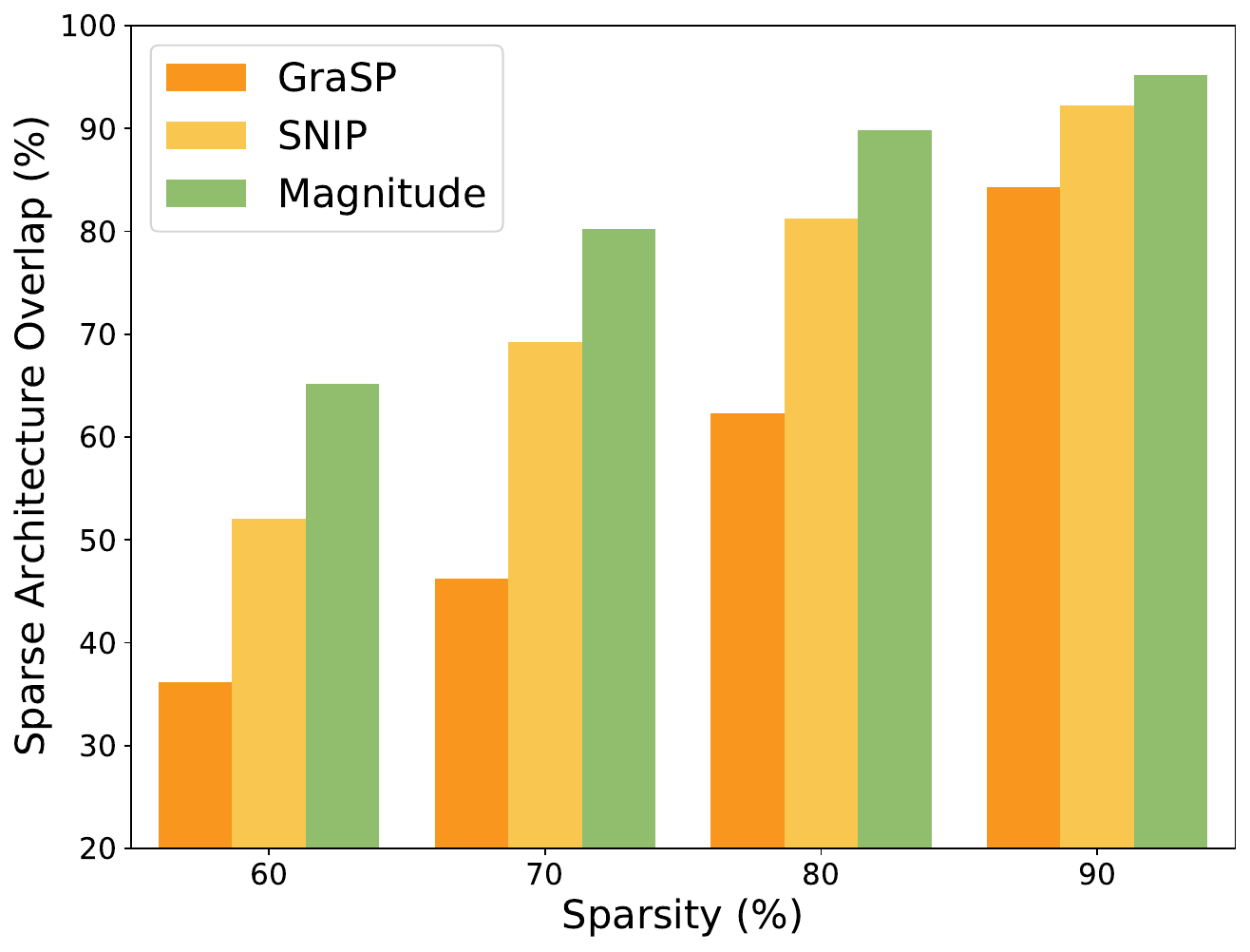}
\end{center}
\caption{\label{overlap}
The comparison of sparse architecture overlap between our searched lottery jackpots and existing pruning criteria without weight training. Experiments are performed using ResNet-50~\cite{he2016deep} on ImageNet~\cite{deng2009imagenet}. 
}
\end{figure}

In Fig.\,\ref{fig:without_retrain}, we show the performance of the pruned models by the above pruning criteria without weight training. As can be observed, our lottery tickets outperform the pruned models by a large margin as the network sparsity goes up. Further, Fig.\,\ref{overlap} indicates a large overlap ratio between the masks of existing methods $\hat{\mathbf{m}}$ and our searched result $h(\bar{\mathbf{m}}$). Among them, the magnitude-based pruning takes the top position. For example, the overlap between magnitude-based mask and our lottery jackpot is more than 95\% when removing around 90\% parameters of ResNet-50~\cite{he2016deep}. This discovery indicates that a very small portion of the pruned mask from existing pruning criteria needs to be corrected to fit our searched mask without the necessity of a time-consuming weight training process in existing work.

Thus, we propose to leverage the importance score $\mathbf{s}$ from existing pruning criteria as a warm-up initialization of our relaxed mask $\mathbf{\bar{m}}$ for searching the lottery jackpots. The relaxed mask $\mathbf{\bar{m}}$ is initialized as:
%
%
\begin{equation}\label{ini_func}
\mathbf{\bar{m}}^i = \left\{ \begin{array}{ll} 
 \eta, \; \textrm{if $\mathbf{s}^i$ in the top-$p\%$ smallest of $\mathbf{s}$,}\\
 1, \; \textrm{otherwise,}
  \end{array} \right.
\end{equation}
where $\eta = 0.99$ in this paper. The rationale behind this is that closer initial mask values of pruned weights to preserved counterparts indicate smaller distances to overcome the remaining non-overlaps between the initialized mask and lottery jackpot, which leads to a fast convergence as we quantitatively demonstrate in Sec.\;\ref{ablation}.

Consequently, the initialized binary mask $h(\bar{\mathbf{m}})$ generated by Eq.\,(\ref{h_func}) is the same as the mask $\mathbf{\hat{m}}$ from existing pruning criteria.
Overall, our target of finding the lottery jackpots can be reformulated as:
\begin{equation}\label{eq:target}
\begin{split}
    \min_{\mathbf{\bar{m}}} \; \mathcal{L} \big(h(\mathbf{\bar{m}}) \odot \tilde{\mathbf{w}} \; ; \mathcal{D}\big),  \;\;\emph{s.t.} \;\; 1 - \frac{\left\| h(\mathbf{\bar{m}}) \right\|_0}{k} \geq p.
\end{split}
\end{equation}
%

\begin{figure*}[!t]
\centering
\begin{subfigure}[t]{0.33\textwidth}
        \centering
        \includegraphics[width=\textwidth]{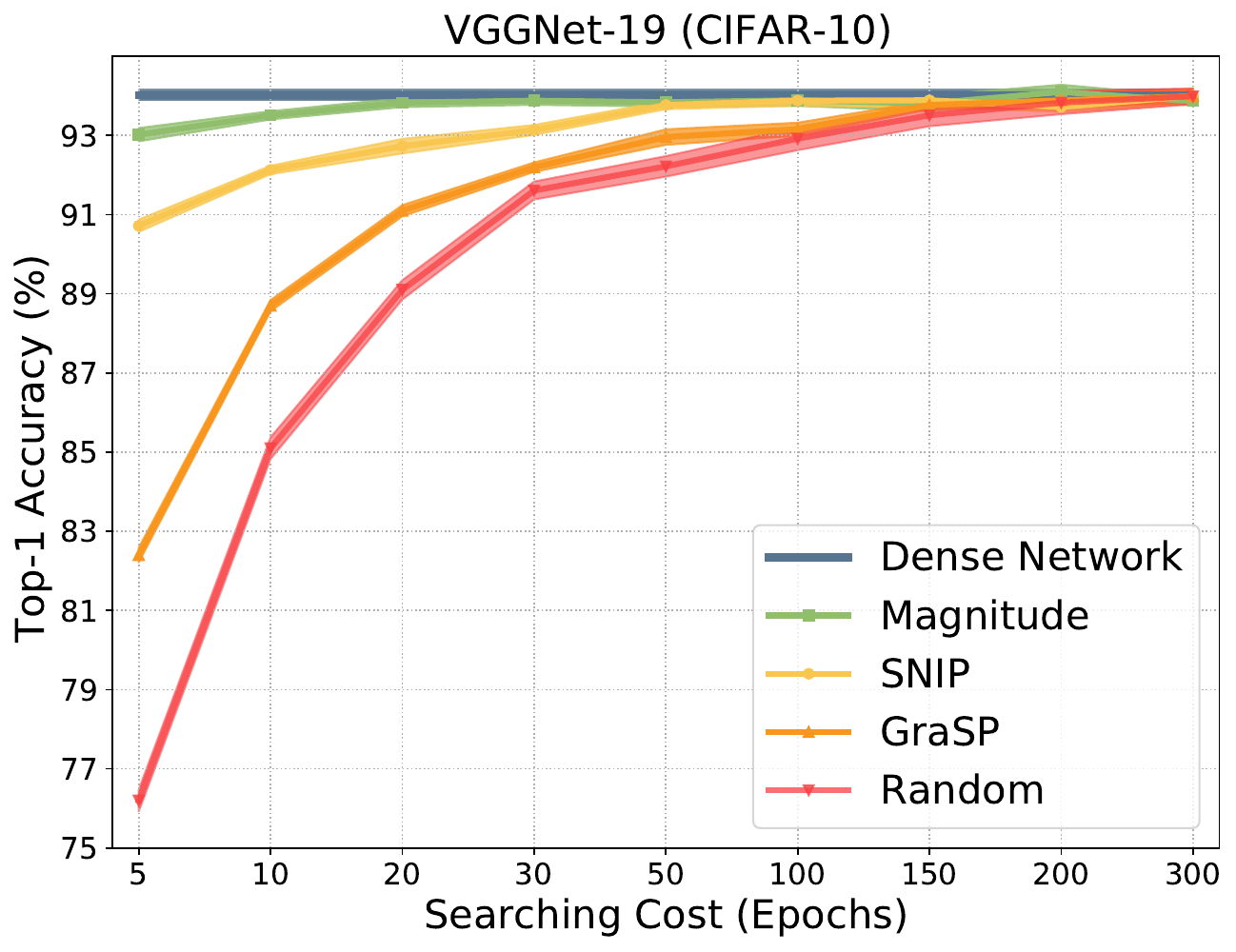}\\
 \end{subfigure}
\begin{subfigure}[t]{0.33\textwidth}
        \centering
        \includegraphics[width=\textwidth]{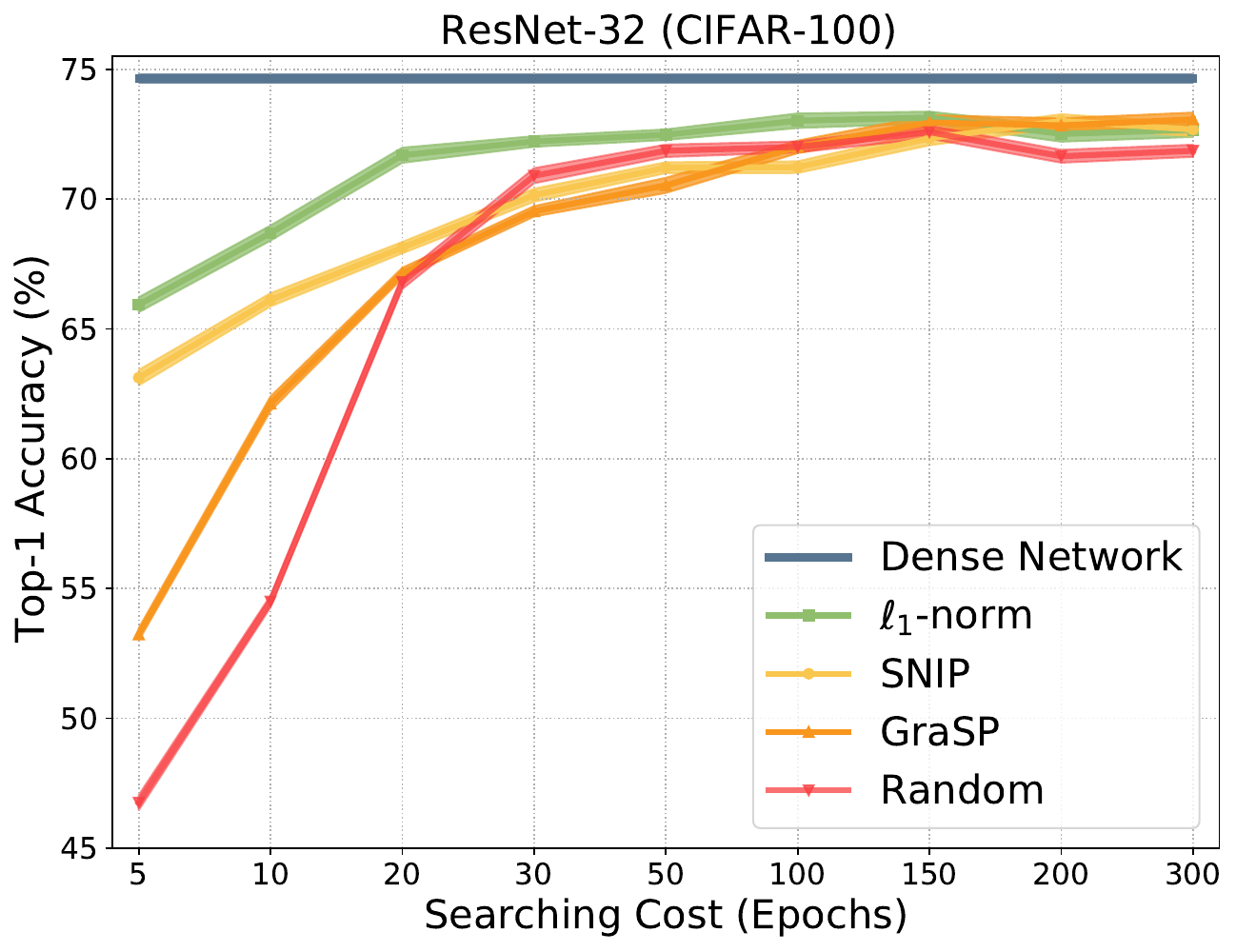}\\
\end{subfigure}
\begin{subfigure}[t]{0.33\textwidth}
        \centering
        \includegraphics[width=\textwidth]{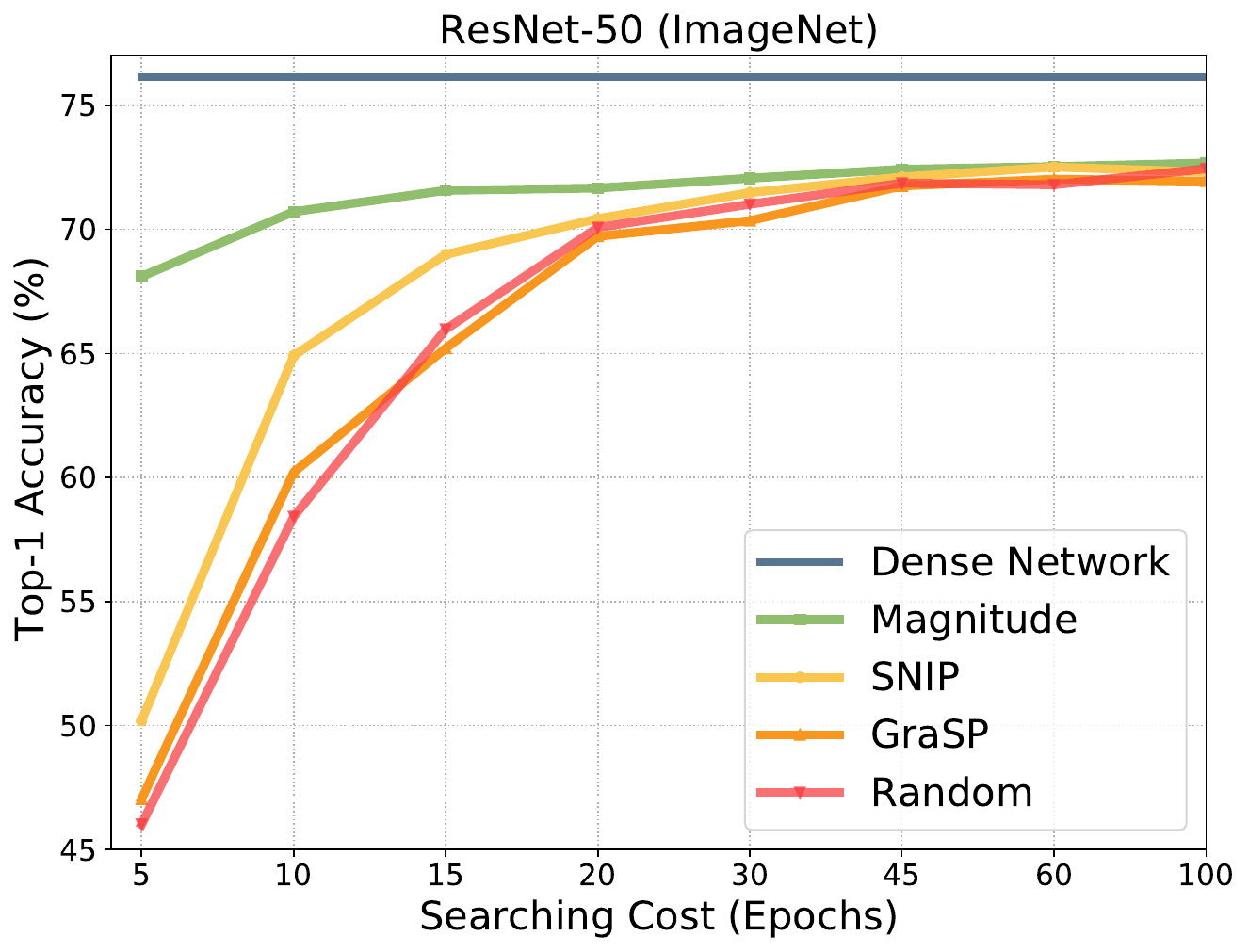}\\
\end{subfigure}
\centering
\caption{Convergence ability of our lottery jackpots using the masks of existing pruning criteria as a warm-up initialization for the weight searching in different networks and benchmarks (90\% sparsity). As can be seen, with the mask from the magnitude pruning as the initialization, our weight searching can easily get convergence with less computation cost.}
\label{fig4}
\end{figure*}

Then, we conduct experiments using different initialized masks from the pruning criteria mentioned above. As can be seen from Fig.\,\ref{fig4}, different warm-up initializations offer a faster convergence of weight searching in finding our lottery jackpots. Particularly, with magnitude-based mask as the initialization, a high-performing lottery jackpot removing around $90\%$ parameters of VGGNet-19, is located quickly using only $30$ searching epochs. In contrast, it is usually $300$ searching epochs that are used to find out a comparable lottery jackpot with randomly initialized weights. Thus, a significant searching complexity can be reduced when using the magnitude-based mask as the initialization. 
It is easy to understand this phenomenon since the magnitude-based pruning mask results in the highest rate of overlap with our searched mask as shown in Fig.\,\ref{overlap}.
We further study how to improve the search{\color{black}ing} efficiency of lottery jackpots after our mask initialization. Before that, we take an in-depth analysis on the searching instability of the edge-popup algorithm~\cite{ramanujan2020s}, which, as we demonstrate, primarily stems from the weight interdependence in modern neural networks~\cite{ramanujan2020s}.

\subsection{Weight Interdependence}\label{interdependence}

Intuitively, Eq.\,(\ref{eq:target}) minimizes the network loss $\mathcal{L}$ by optimizing the relaxed mask $\mathbf{\bar{m}}$ at each training iteration.
Let the relaxed mask at the $t$-th training iteration be $\mathbf{\bar{m}}_t$, which generates binary masks to prune or preserve weights through Eq.\,(\ref{h_func}).
Since we adopt magnitude-based mask, mask values for preserved weights are larger than those of pruned weights.
Denoting the mask indexes corresponding to the preserved and pruned weights as $\Phi_t$ and $\Psi_t$, we can have the following relationship:
\begin{equation}\label{eq:mask_ueq}
\begin{split}
    min(\mathbf{\bar{m}}^{\Phi_t}_{t}) > max(\mathbf{\bar{m}}^{\Psi_t}_{t}),
\end{split}
\end{equation}
where $min(\cdot)$ and $max(\cdot)$ return the lowest value and highest value, respectively.
Nevertheless, the above inequality might be broken after mask updating at the $t+1$ iteration since some elements in $\mathbf{\bar{m}}^{\Psi_t}_{t+1}$ become larger than some elements in $\mathbf{\bar{m}}^{\Phi_t}_{t+1}$.
%
%
Consequently, these elements in $\mathbf{\bar{m}}^{\Psi_t}_{t+1}$ are revived while these in $\mathbf{\bar{m}}^{\Phi_t}_{t+1}$ are re-pruned at the $t+1$ iteration, which we name as weight swapping in this paper.
Let the indexes of these re-pruned and revived weights be $\Phi_t^* \subseteq  \Phi_t$ and $\Psi_t^* \subseteq \Psi_t$, Proposition 1 reveals that weight swapping decreases the training loss $\mathcal{L}$ under Assumption 1.
%

%
\textbf{Assumption 1}: Let $\Delta \tilde{\mathbf{w}}^i$ denote some weight perturbation on a specific weight $\tilde{\mathbf{w}}^i$ and $\mathcal{L}(\tilde{\mathbf{w}}^i+ \Delta \tilde{\mathbf{w}}^i)$\footnote{The notations of other weights and the observed training set are neglected for simplicity.} denote the new loss after adding the perturbation. Weights $\tilde{\mathbf{w}}^i$ and $\tilde{\mathbf{w}}^j$ independently contribute to the loss function $\mathcal{L}$ and the following constraint satisfies:
\begin{equation}\label{assumption1}
\begin{split}
    &\mathcal{L}(\tilde{\mathbf{w}}^i, \tilde{\mathbf{w}}^j) -  \mathcal{L}(\tilde{\mathbf{w}}^i + \Delta \tilde{\mathbf{w}}^i, \tilde{\mathbf{w}}^j + \Delta \tilde{\mathbf{w}}^j) \\
     &= \mathcal{L}(\tilde{\mathbf{w}}^i, \tilde{\mathbf{w}}^j) - \mathcal{L}(\tilde{\mathbf{w}}^i + \Delta \tilde{\mathbf{w}}^i, \tilde{\mathbf{w}}^j) \\
    &+  \mathcal{L}(\tilde{\mathbf{w}}^i, \tilde{\mathbf{w}}^j) - \mathcal{L}(\tilde{\mathbf{w}}^j + \Delta \tilde{\mathbf{w}}^j, \tilde{\mathbf{w}}^i). 
\end{split}
\end{equation}

\textbf{Proposition 2}: \textit{For $\forall i \in \Phi^*_t$, $\forall j \in \Psi^*_t$, the following inequality can always be satisfied:}
\begin{equation}
         \mathcal{L}\big(h(\mathbf{\bar{m}}^i_{t+1})\tilde{\mathbf{w}}^i, h(\mathbf{\bar{m}}^j_{t+1})\tilde{\mathbf{w}}^j\big) <  \mathcal{L}\big(h(\mathbf{\bar{m}}^i_{t})\tilde{\mathbf{w}}^i, h(\mathbf{\bar{m}}^j_{t})\tilde{\mathbf{w}}^j\big).
\end{equation}
%

\textbf{Proof}.
$\mathcal{ L}\big(h(\mathbf{\bar{m}}_{t+1}^i) \tilde{\mathbf{w}}^i)$ indicates the loss after removing the $\tilde{\mathbf{w}}^i$ and it can be approximated according to the first-order Taylor series in the following:
\begin{equation}\label{eq:prune}
\begin{split}
    &\mathcal{L}\big(h(\mathbf{\bar{m}}_{t+1}^i) \tilde{\mathbf{w}}^i\big) \\
   \approx &\mathcal{L}\big(h(\mathbf{\bar{m}}_{t}^i)\tilde{\mathbf{w}}^i\big)- \frac{\partial \mathcal{L}}{\partial \big(h(\mathbf{\bar{m}}_t^i) \tilde{\mathbf{w}}^i \big)} \big(h(\mathbf{\bar{m}}_{t}^i)\tilde{\mathbf{w}}^i- h(\mathbf{\bar{m}}_{t+1}^i)\tilde{\mathbf{w}}^i\big) \\
   = & \mathcal{L}\big(h(\mathbf{\bar{m}}_{t}^i)\tilde{\mathbf{w}}^i\big)- \frac{\partial \mathcal{L}}{\partial \big(h(\mathbf{\bar{m}}_t^i) \tilde{\mathbf{w}}^i \big)} \tilde{\mathbf{w}}^i.
\end{split}
\end{equation}

And $\mathcal{L}\big(h(\mathbf{\bar{m}}_{t+1}^j) \tilde{\mathbf{w}}^j)$ denotes the loss after reviving the $\tilde{\mathbf{w}}^j$. Similarly, it can be approximated as:
\begin{equation}\label{eq:revive}
\begin{split}
    & \quad\mathcal{L}\big(h(\mathbf{\bar{m}}_{t+1}^j) \tilde{\mathbf{w}}^j\big) \\
   &\approx \mathcal{L}\big(h(\mathbf{\bar{m}}_{t}^j)\tilde{\mathbf{w}}^j\big)- \frac{\partial \mathcal{L}}{\partial \big(h(\mathbf{\bar{m}}_t^j) \tilde{\mathbf{w}}^j\big)} \big(h(\mathbf{\bar{m}}_{t}^j)\tilde{\mathbf{w}}^j- h(\mathbf{\bar{m}}_{t+1}^j)\tilde{\mathbf{w}}^j\big) \\
   &= \mathcal{L}\big(h(\mathbf{\bar{m}}_{t}^j)\tilde{\mathbf{w}}^j\big)+ \frac{\partial \mathcal{L}}{\partial \big(h(\mathbf{\bar{m}}_t^j) \tilde{\mathbf{w}}^j \big)} \tilde{\mathbf{w}}^j.
\end{split}
\end{equation}

Combining the above two equalities yields:
\begin{equation}\label{eq:losschange_1}
\begin{split}
    &\mathcal{L}\big(h(\mathbf{\bar{m}}_{t+1}^j) \tilde{\mathbf{w}}^j) - \mathcal{L}\big(h(\mathbf{\bar{m}}_{t}^j)\tilde{\mathbf{w}}^j\big)
    \\&
    + \mathcal{L}\big(h(\mathbf{\bar{m}}_{t+1}^i) \tilde{\mathbf{w}}^i) - \mathcal{L}\big(h(\mathbf{\bar{m}}_{t}^i)\tilde{\mathbf{w}}^i\big)\\
    &= \frac{\partial \mathcal{L}}{\partial \big(h(\mathbf{\bar{m}}_t^j) \tilde{\mathbf{w}}^j \big)} \tilde{\mathbf{w}}^j - \frac{\partial \mathcal{L}}{\partial \big(h(\mathbf{\bar{m}}_t^i) \tilde{\mathbf{w}}^i \big)} \tilde{\mathbf{w}}^i.
\end{split}
\end{equation}

Considering Assumption 3 we have:
\begin{equation}\label{eq:final-1}
\begin{split}
    &\mathcal{L}\big(h(\mathbf{\bar{m}}^i_{t+1})\tilde{\mathbf{w}}^i,  h(\mathbf{\bar{m}}^j_{t+1})\tilde{\mathbf{w}}^j\big) -  \mathcal{L}\big(h(\mathbf{\bar{m}}^i_{t})\tilde{\mathbf{w}}^i, h(\mathbf{\bar{m}}^j_{t})\tilde{\mathbf{w}}^j\big)\\
    &=\frac{\partial \mathcal{L}}{\partial \big(h(\mathbf{\bar{m}}_t^j) \tilde{\mathbf{w}}^j \big)} \tilde{\mathbf{w}}^j - \frac{\partial \mathcal{L}}{\partial \big(h(\mathbf{\bar{m}}_t^i) \tilde{\mathbf{w}}^i \big)} \tilde{\mathbf{w}}^i.
\end{split}
\end{equation}

By Eq.\,(\ref{eq:mask_ueq}) we have $\mathbf{\bar{m}}^i_{t} > \mathbf{\bar{m}}^j_{t}$, and given $i \in \Phi_t^*$, $j \in \Psi_t^*$ we obtain $\mathbf{\bar{m}}^j_{t+1} > \mathbf{\bar{m}}^i_{t+1}$. Combining these two inequalities therefore results in:
\begin{equation}\label{eq:mask_uneq_2}
 \mathbf{\bar{m}}^i_{t} - \mathbf{\bar{m}}^i_{t+1} > \mathbf{\bar{m}}^j_{t} - \mathbf{\bar{m}}^j_{t+1}.
\end{equation}

Looking back to Eq.\,(\ref{mask_g}), the updated state of $\mathbf{\bar{m}}$ from the $t$-th iteration to the ($t+1$)-th can derived as:
\begin{equation}\label{m_t_iter}
\begin{split}
    \mathbf{\bar{m}}_{t+1} &= \mathbf{\bar{m}}_t - \frac{\partial \mathcal{L}}{\partial \mathbf{\bar{m}}_t} \\
    &= \mathbf{\bar{m}}_t - \frac{\mathcal{\partial L}}{\partial \big(h(\mathbf{\bar{m}}_t) \odot \tilde{\mathbf{w}} \big)} \frac{\partial \big(h(\mathbf{\bar{m}}_t) \odot \tilde{\mathbf{w}} \big)}{\partial \mathbf{\bar{m}}_t}\\
    & = \mathbf{\bar{m}}_t -  \frac{\mathcal{\partial L}}{\partial \big(h(\mathbf{\bar{m}}_t) \odot \tilde{\mathbf{w}} \big)} \frac{\partial h(\mathbf{\bar{m}}_t) }{\partial \mathbf{\bar{m}}_t} \tilde{\mathbf{w}}\\
    &\approx \mathbf{\bar{m}}_t - \frac{\mathcal{\partial L}}{\partial \big(h(\mathbf{\bar{m}}_t) \odot \tilde{\mathbf{w}} \big)}  \tilde{\mathbf{w}},
\end{split}
\end{equation}
where learning rate and optimizer items like weight decay are neglected for ease of representation.
By taking this derivation into Eq.\,(\ref{eq:mask_uneq_2}), we obtain:
\begin{equation}\label{eq:update_uneq}
 \frac{\mathcal{\partial L}}{\partial \big(h(\mathbf{\bar{m}}_t^i) \tilde{\mathbf{w}}^i \big)}  \tilde{\mathbf{w}}^i > \frac{\mathcal{\partial L}}{\partial \big(h(\mathbf{\bar{m}}_t^j) \tilde{\mathbf{w}}^j \big)}  \tilde{\mathbf{w}}^j.
\end{equation}

This inequality and Eq.\,(\ref{eq:final-1}) lead to:
\begin{equation}\label{eq:final}
\begin{split}
    & \mathcal{L}\big(h(\mathbf{\bar{m}}^i_{t+1})\tilde{\mathbf{w}}^i,  h(\mathbf{\bar{m}}^j_{t+1})\tilde{\mathbf{w}}^j\big) -  \mathcal{L}\big(h(\mathbf{\bar{m}}^i_{t})\tilde{\mathbf{w}}^i, h(\mathbf{\bar{m}}^j_{t})\tilde{\mathbf{w}}^j\big) \\
    &= \frac{\partial \mathcal{L}}{\partial \big(h(\mathbf{\bar{m}}_t^j) \tilde{\mathbf{w}}^j \big)} \tilde{\mathbf{w}}^j - \frac{\partial \mathcal{L}}{\partial \big(h(\mathbf{\bar{m}}_t^i) \tilde{\mathbf{w}}^i \big)} \tilde{\mathbf{w}}^i <0.
\end{split}
\end{equation}

Consequently, we have:
\begin{equation}
         \mathcal{L}\big(h(\mathbf{\bar{m}}^i_{t+1})\tilde{\mathbf{w}}^i, h(\mathbf{\bar{m}}^j_{t+1})\tilde{\mathbf{w}}^j\big) <  \mathcal{L}\big(h(\mathbf{\bar{m}}^i_{t})\tilde{\mathbf{w}}^i, h(\mathbf{\bar{m}}^j_{t})\tilde{\mathbf{w}}^j\big),
\end{equation}
which finally complete our proof of Proposition 1. $\hfill\blacksquare$

\begin{figure}[!t]
\centering
\begin{center}
\includegraphics[height=0.62\linewidth]{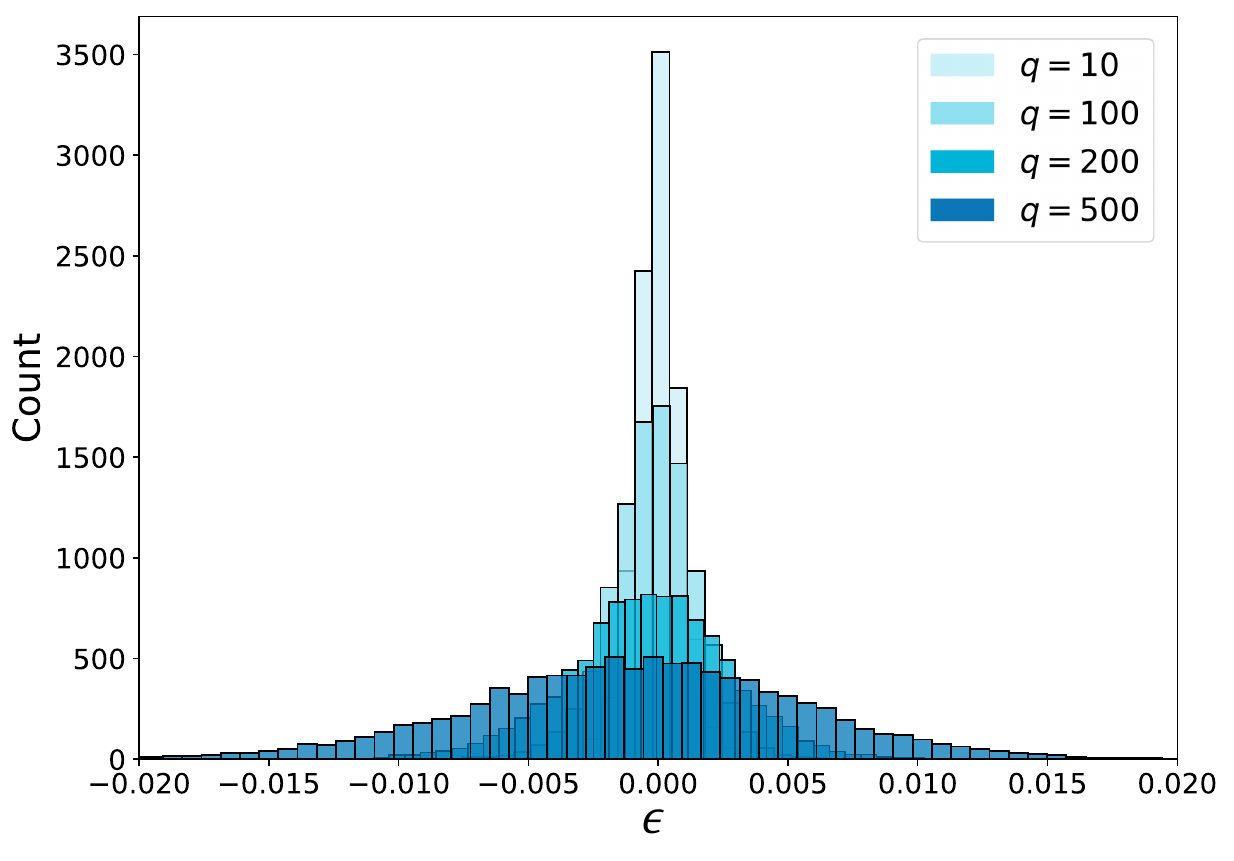}
\end{center}
\centering
\caption{\label{eplsilon}The distortion error $\epsilon$ caused by the interdependence among weights \emph{v.s.} quantities of weight swapping $q$. Experiments are performed using ResNet-32~\cite{he2016deep}.}
\end{figure}

\begin{figure*}[!t]
\centering
\begin{subfigure}[t]{0.33\textwidth}
        \centering
        \includegraphics[width=\textwidth]{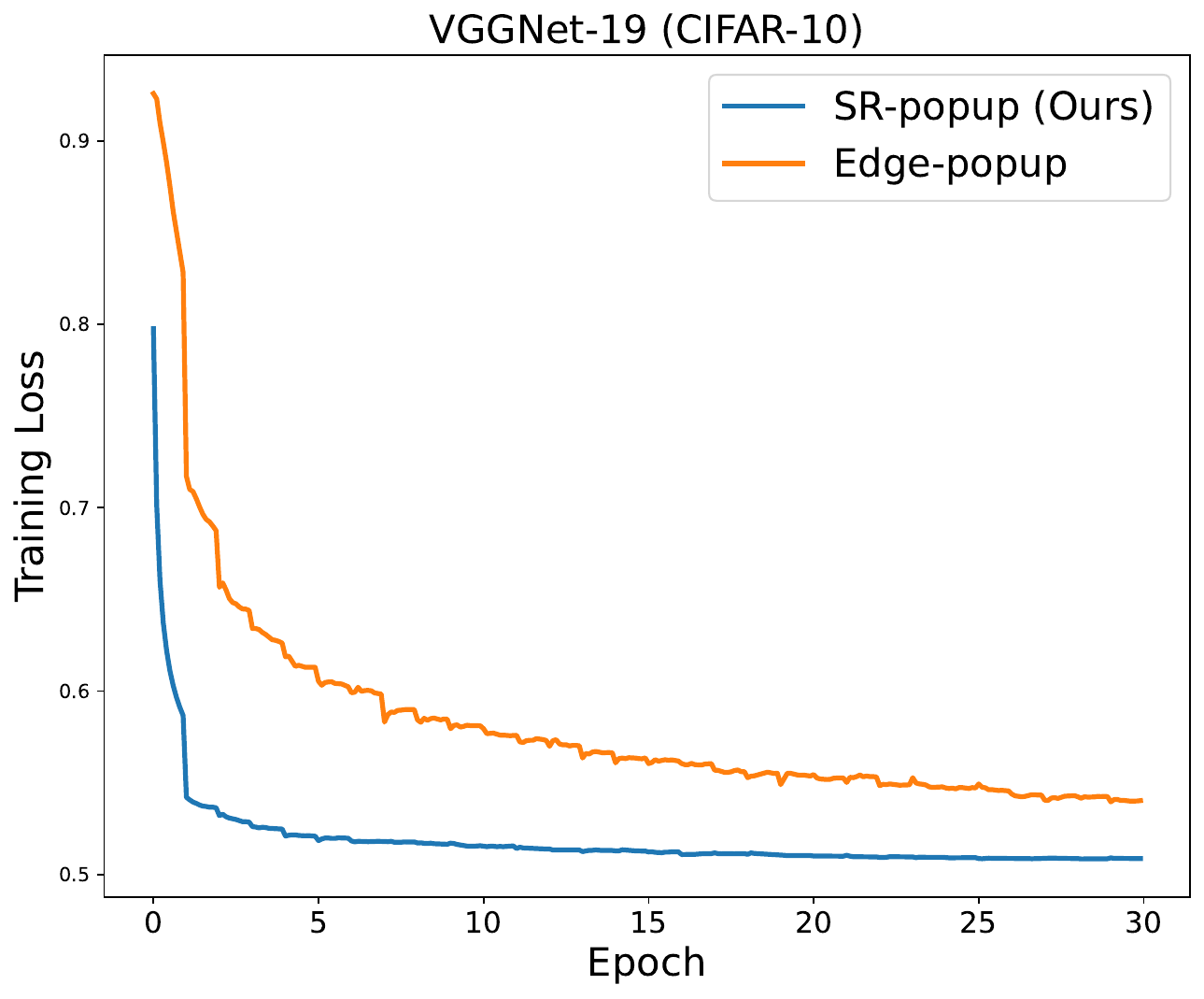}\\
 \end{subfigure}
\begin{subfigure}[t]{0.33\textwidth}
        \centering
        \includegraphics[width=\textwidth]{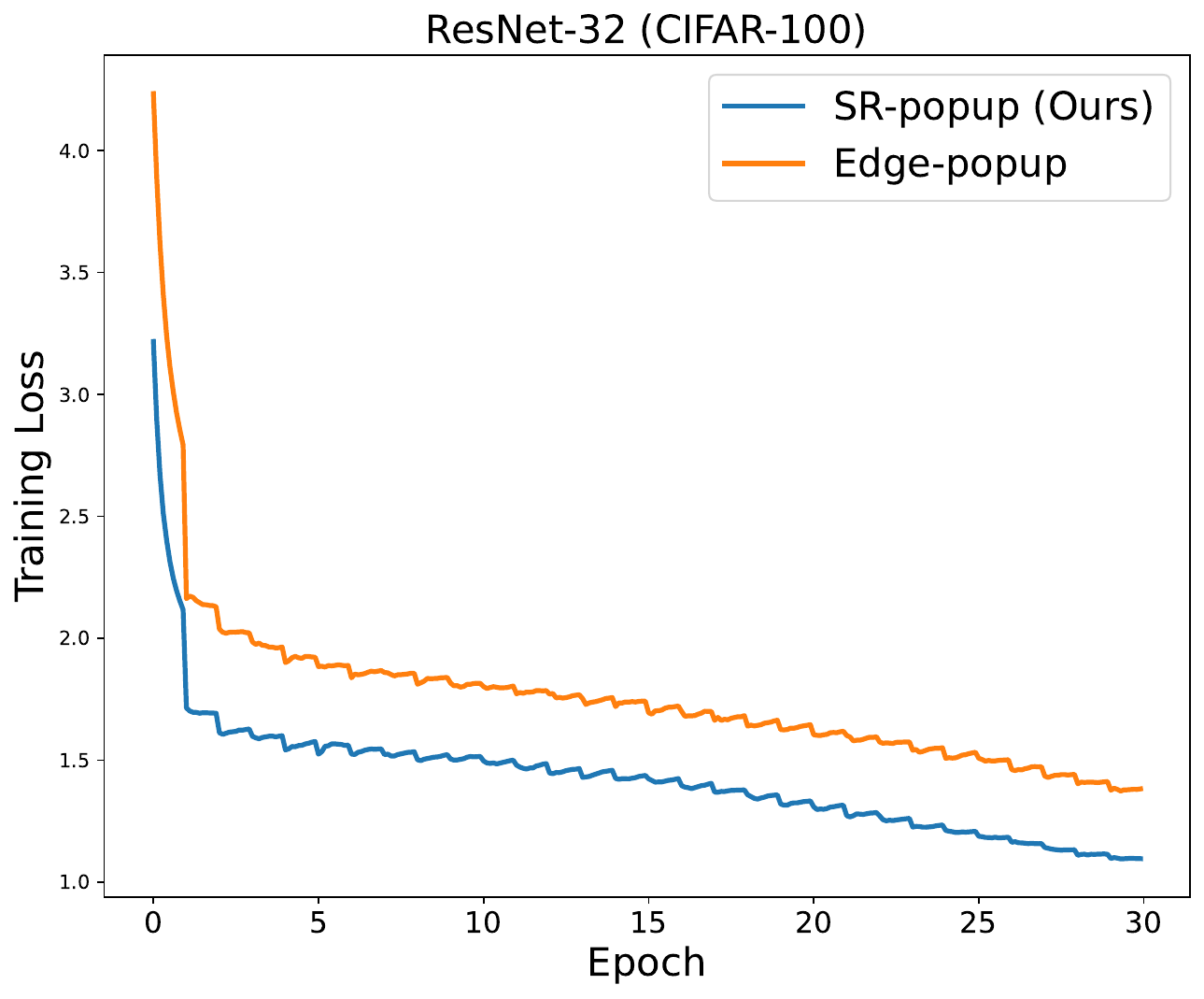}\\
\end{subfigure}
\begin{subfigure}[t]{0.33\textwidth}
        \centering
        \includegraphics[width=\textwidth]{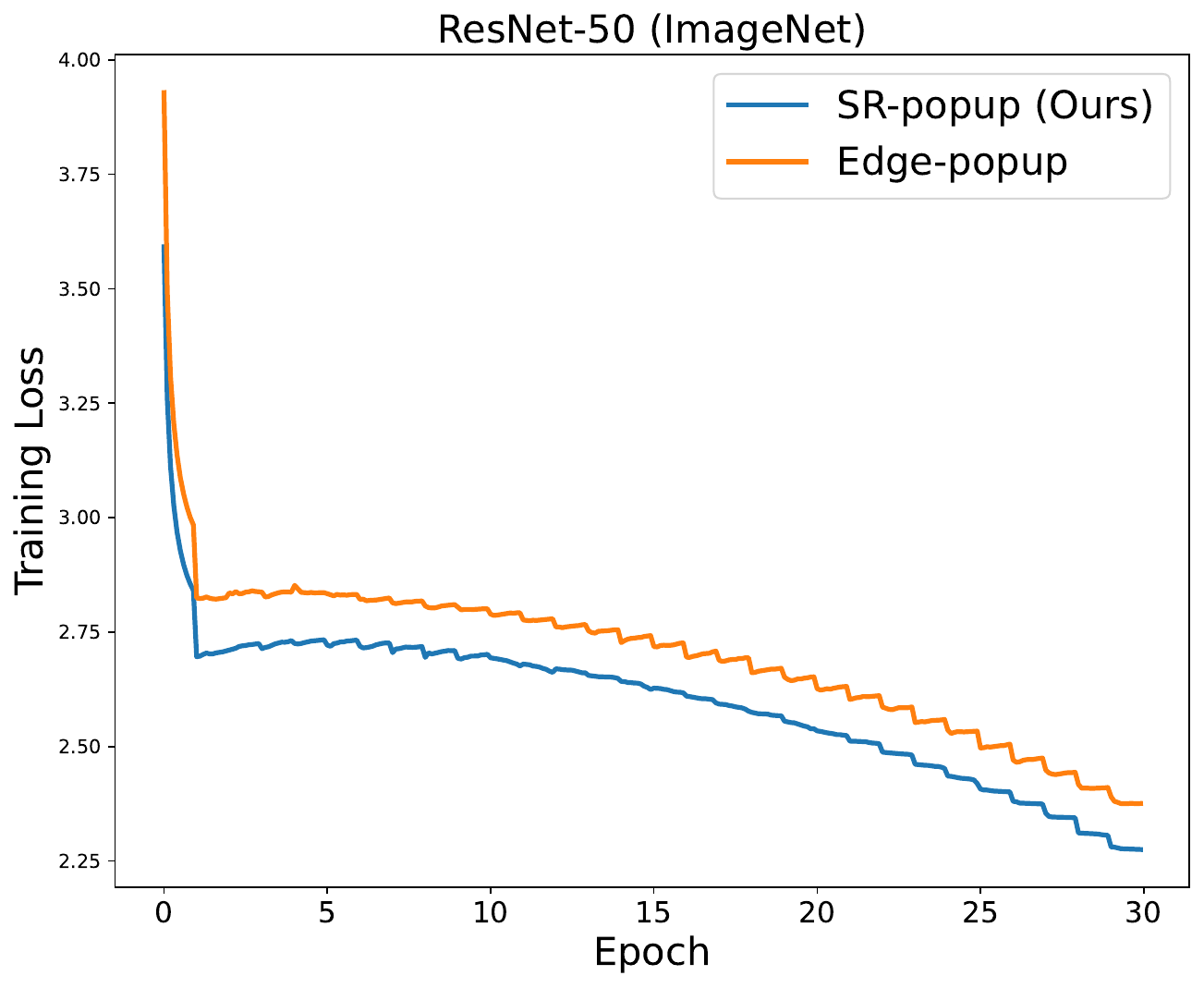}\\
\end{subfigure}
\centering
\caption{Loss curve of our proposed SR-popup and the Edge-popup algorithms for searching lottery jackpots in different networks and benchmarks (90\% sparsity). As can be seen, SR-popup exhibits higher stability and convergence speed.}
\label{fig:loss_curve}
\end{figure*}

So far, we have proved that Proposition 1 guarantees the loss drop for weight swapping between $\tilde{\mathbf{w}}^i$ and $\tilde{\mathbf{w}}^j$ under Assumption 1, and the expected loss drop denoted by $\Delta \mathcal{L}(\tilde{\mathbf{w}}^i, \tilde{\mathbf{w}}^j)$ is:
\begin{equation}\label{eq:expect_decrease}
    \Delta \mathcal{L}(\tilde{\mathbf{w}}^i, \tilde{\mathbf{w}}^j) = \frac{\partial \mathcal{L}}{\partial \big(h(\mathbf{\bar{m}}_t^j) \tilde{\mathbf{w}}^j \big)} \tilde{\mathbf{w}}^j - \frac{\partial \mathcal{L}}{\partial \big(h(\mathbf{\bar{m}}_t^i) \tilde{\mathbf{w}}^i \big)} \tilde{\mathbf{w}}^i.
\end{equation}

Unfortunately, Assumption 1 does not hold always in modern neural networks due to the existence of Batch normalization layers, Softmax function,~\emph{etc}.
In contrast, the interdependence frequently stems among different weights as discussed in~\cite{wang2020picking, nagel2020up, zhang2022optimizing}.
Under this case, Eq.\,(\ref{assumption1}) is indeed presented as:
\begin{equation}\label{assumption1_noram}
\begin{split}
    & \mathcal{L}(\tilde{\mathbf{w}}^i, \tilde{\mathbf{w}}^j) -  \mathcal{L}(\tilde{\mathbf{w}}^i + \Delta \tilde{\mathbf{w}}^i, \tilde{\mathbf{w}}^j + \Delta \tilde{\mathbf{w}}^j) \\
    & = \mathcal{L}(\tilde{\mathbf{w}}^i, \tilde{\mathbf{w}}^j) - \mathcal{L}(\tilde{\mathbf{w}}^i + \Delta \tilde{\mathbf{w}}^i, \tilde{\mathbf{w}}^j) \\
    &+  \mathcal{L}(\tilde{\mathbf{w}}^i, \tilde{\mathbf{w}}^j) - \mathcal{L}(\tilde{\mathbf{w}}^j + \Delta \tilde{\mathbf{w}}^j, \tilde{\mathbf{w}}^i) + \epsilon,
\end{split}
\end{equation}
where $\epsilon \in \mathbb{R}$ is the distortion error caused by the dependency of weights.
Repeating the deduction procedures for Proposition 1, we obtain:
\begin{equation}\label{eq:final*}
\begin{split}
 &\mathcal{L}\big(h(\mathbf{\bar{m}}^i_{t+1})\tilde{\mathbf{w}}^i,  h(\mathbf{\bar{m}}^j_{t+1})\tilde{\mathbf{w}}^j\big) -  \mathcal{L}\big(h(\mathbf{\bar{m}}^i_{t})\tilde{\mathbf{w}}^i, h(\mathbf{\bar{m}}^j_{t})\tilde{\mathbf{w}}^j\big)  
 \\ &
 = \underbrace{\frac{\partial \mathcal{L}}{\partial \big(h(\mathbf{\bar{m}}_t^j) \tilde{\mathbf{w}}^j \big)} \tilde{\mathbf{w}}^j - \frac{\partial \mathcal{L}}{\partial \big(h(\mathbf{\bar{m}}_t^i) \tilde{\mathbf{w}}^i \big)} \tilde{\mathbf{w}}^i}_{\Delta \mathcal{L}(\tilde{\mathbf{w}}^i, \tilde{\mathbf{w}}^j) \;\text{in Eq.\,(\ref{eq:expect_decrease})}} + \epsilon.
\end{split}
\end{equation}

Hence, the actual loss drop for the weight swapping falls in $ \Delta \mathcal{L}(\tilde{\mathbf{w}}^i, \tilde{\mathbf{w}}^j) + \epsilon$.
It is easy to know that Eq.\,(\ref{eq:final}) always holds and the loss still decreases from weight swapping if the following constraint satisfies:
\begin{equation}\label{constraint}
\epsilon < -\Delta \mathcal{L}(\tilde{\mathbf{w}}^i, \tilde{\mathbf{w}}^j).
\end{equation}

Otherwise, Eq.\,(\ref{eq:final}) breaks and the loss even becomes larger after updating, in particular to frequent weight swapping. Fig.\;\ref{eplsilon} displays quantitative distributions of $\epsilon$ by randomly performing weight swapping in different quantities of swapped weights for ResNet-32~\cite{he2016deep}. We can see that the variance of the distortion error $\epsilon$ increases proportionally to the number of swapped weights, destroying the above constraint.
%
%
As a result, the searching stability of the edge-popup algorithm is bounded by such a distortion error stemming from the weight interdependence.

{\subsection{Short Restriction Popup}\label{popup}

%
%
To solve the above issue, on the premise of mask initialization introduced in Sec.\,\ref{fast}, we further propose a short restriction popup to improve the searching efficiency for lottery jackpots.
Our key motivation is to encourage weight swapping if we observe a small value of expected loss drop $\Delta \mathcal{L}(\tilde{\mathbf{w}}^i, \tilde{\mathbf{w}}^j)$ which is robust to the distortion error according to Eq.\,(\ref{constraint}); otherwise, the weight swapping is going to be prevented. Consequently, the continuously decreasing loss results in a stable training convergence.

%

%

%
Concretely, we first obtain the indexes of preserved and pruned weights from the initialized mask $\mathbf{\bar{m}}_0$ in Eq.\,(\ref{ini_func}):
\begin{equation}\label{psi_0}
 \begin{array}{ll} 
\Phi_0 = \{i \; | \; \mathbf{\bar{m}}^i=1, i = 1, 2, ..., k\}, \\
\Psi_0 = \{i \; | \; \mathbf{\bar{m}}^i=\eta, i = 1, 2, ..., k\},
\end{array}
\end{equation}
where $k$ is the weight size as recalled in Sec.\,\ref{preliminary}. In the $t$-th forward propagation, we obtain the binary mask by:
\begin{equation}\label{h_func_1}
h^*(\mathbf{\bar{m}}^i) = \left\{ \begin{array}{ll} 
 1, \; \textrm{if } i \in \Phi_t,\\
 0, \; \textrm{otherwise,}
  \end{array} \right.
\end{equation}
The backward propagation is the same as Eq.\,(\ref{mask_g}).
After the $t$-th updating, the indexes of re-pruned and re-revived weights $\Phi^*_t$ and $\Psi^*_t$ are obtained by: 
\begin{equation}\label{psi*}
 \begin{array}{ll}
 \Phi^*_t = \{i \; | \; \mathbf{\bar{m}}^i_{t+1} \leq TopK(\mathbf{\bar{m}}_{t+1}, \left\| \Psi_t \right\|_0), i \in \Phi_t  \}, \\
\Psi^*_t = \{i \; | \; \mathbf{\bar{m}}^i_{t+1} > TopK(\mathbf{\bar{m}}_{t+1}, \left\| \Psi_t \right\|_0), i \in \Psi_t  \},
\end{array}
\end{equation}
where $TopK(\mathbf{v}, K)$ returns the $K$-$th$ largest value within vector $\mathbf{v}$. 
Then, we gradually decrease the weight swapping number 
 $q_t$ between $\mathbf{\bar{m}}^{\Psi^*_t}_{t}$ and $\mathbf{\bar{m}}^{\Phi^*_t}_{t}$ as:
\begin{equation}\label{eq:schedule}
    q_t = \lceil \left\|\Psi^*_t\right\|_0 (1- \frac{t}{t_f})^4\rceil,
\end{equation}
where $\lceil \cdot \rceil$ is the ceiling operation and $t_f$ is the total training iterations.
To explain, we encourage more weight swapping to achieve enough exploration at the early stage of training, and then gradually limit the number of weight swapping to achieve efficient convergence by more loss drops.
We choose to swap the weights of top-$q_t$ smallest $\mathbf{\bar{m}}^{\Phi^*_t}$ and weights of top-$q_t$ largest $\mathbf{\bar{m}}^{\Psi^*_t}$.
Consequently, the mask indexes corresponding to the preserved weights and the pruned weights at iteration $t+1$ are obtained as:
\begin{equation}\label{index_update}
 \begin{array}{ll}
 \Phi_{t+1} = \Phi_{t} \setminus ArgBotK(\mathbf{\bar{m}}_{t+1}^{\Phi_t^*}, q_t)\cup ArgTopK(\mathbf{\bar{m}}_{t+1}^{\Psi_t^*}, q_t),\\
\Psi_{t+1} = \Psi_{t} \setminus ArgTopK(\mathbf{\bar{m}}_{t+1}^{\Psi_t^*}, q_t) \cup ArgBotK(\mathbf{\bar{m}}_{t+1}^{\Phi_t^*}, q_t),
\end{array}
\end{equation}
where $ArgTopK(\mathbf{v}, K)$, $ArgBotK(\mathbf{v}, K)$ returns the top-$K$ largest and smallest values within vector $\mathbf{v}$.

\begin{algorithm}[!t]

\caption{Short Restriction Popup for Locating Lottery Jackpots}\label{alg1}
\LinesNumbered
\KwIn{Pre-trained weights $\tilde{\mathbf{w}}$, pruning rate $p$, searching iteration $t_f$.}
Initialize the relaxed mask $\mathbf{\bar{m}}_0$ via Eq.\,(\ref{ini_func}).\\
Obtain preserved and pruned weight indexes $\Phi_0$ and $\Psi_0$ via Eq.\,(\ref{psi_0}). \\
\For{t = 1 $\rightarrow$ $t_f$}{
   Forward propagation via Eq.\,(\ref{h_func_1}).  \\
   Backward propagation via Eq.\,(\ref{mask_g}). \\
   Update $\mathbf{\bar{m}}$ using SGD optimizer. \\
   Conduct weight swapping via Eq.\,(\ref{index_update}). \\
}

Return the compressed model $h^*(\mathbf{\bar{m}}) \odot \tilde{\mathbf{w}}$.
\end{algorithm}

Our proposed short restriction pop-up, termed SR-popup for searching lottery jackpots, is listed in Alg.\,\ref{alg1}.
Fig.\,\ref{fig:loss_curve} shows the loss curves of the SR-popup and Edge-popup when searching lottery jackpots across various datasets.
By escaping weight swapping that is not robust to the distortion error, SR-popup effectively reduces the loss oscillation and leads to a more stable searching process.

\section{Experiments}\label{experiment}

\subsection{Settings}\label{setting}

\textbf{Datasets}. We consider representative benchmarks with different scales for image classification. For the small-scale dataset, we choose the CIFAR-10 and CIFAR-100 datasets~\cite{krizhevsky2009learning}. 
CIFAR-10 contains 60,000 32$\times$32 color images from 10 different classes, with 6,000 images in each class. The CIFAR-100 dataset is similar to the CIFAR-10, except that it has 100 classes, each of which contains 600 images. 
For the large-scale dataset, we choose the challenging ImageNet-1K~\cite{deng2009imagenet} that has over 1.2 million images for training and 50,000 validation images with 1,000 categories. 

\textbf{Networks}. For CIFAR-10 and CIFAR-100, we find lottery jackpots in the classic VGGNet-19~\cite{simonyan2015very} and ResNet-32~\cite{he2016deep}. 
Following previous studies~\cite{wang2020picking,hayou2020pruning}, we double the filter numbers of each convolution layer of ResNet-32 to make it suitable to fit the dataset.
For the ImageNet dataset, we demonstrate the efficacy of lottery jackpots for pruning ResNet-50~\cite{he2016deep}, which serves as a typical backbone for network pruning community.
Besides, we conduct experiments to find lottery jackpots for pruning very light-weight networks including MobileNet-V1~\cite{howard2017mobilenets}, EfficientNet-B4~\cite{tan2019efficientnet}. 
%
%

\begin{table}[!t]
\setlength{\tabcolsep}{0.5em}
\setlength{\abovecaptionskip}{2pt}
\caption{\label{cifar10} Comparison with existing weight training methods for pruning VGGNet-19~\cite{simonyan2015very} and ResNet-32~\cite{he2016deep} on CIFAR-10~\cite{krizhevsky2009learning}. We report top-1 accuracy (\%), training cost (T) and searching cost (S).}
\centering
\resizebox{\columnwidth}{!}{
\begin{tabular}{c|cc|c}
\toprule
Pruning Rate& \quad\quad90\%\quad\quad  & \quad\quad95\%\quad\quad\quad\quad  & Epoch\\
\midrule
VGGNet-19 &  94.23 & - & -  \\
SET~\cite{mocanu2018scalable} & 92.46 & 91.73 & 160 (T) \\
Deep-R~\cite{bellec2017deep} & 90.81 & 89.59   & 160 (T) \\
SNIP~\cite{lee2018snip} &  93.65 & 93.43 & 160(T)\\
GraSP~\cite{wang2020picking} & 93.01 & 92.82   & 160 (T) \\
LT~\cite{frankle2018lottery} &  93.66 & 93.30  & 160 (T) \\
OBD~\cite{lecun1989optimal} & 93.74& 93.58 & 160 (T)\\
Jackpot (Ours)  & 93.70$\pm$0.12 & 93.54$\pm$0.17  & \textbf{10 (S)}    \\
\textbf{Jackpot (Ours)}  & \textbf{93.86$\pm$0.09} & \textbf{93.67$\pm$0.12} & 30 (S)   \\
\midrule                                
ResNet-32    &      94.62    &  -      &       -       \\
SET~\cite{mocanu2018scalable} &  92.30 & 90.76 &  160 (T) \\
Deep-R~\cite{bellec2017deep} & 91.62 & 89.84  & 160 (T) \\
SNIP~\cite{lee2018snip} & 92.59 & 91.01&  160 (T) \\
GraSP~\cite{wang2020picking} & 92.79 & 91.80   &  160 (T) \\ 
LT~\cite{frankle2018lottery} & 92.61 & 91.37   & 160 (T) \\
OBD~\cite{lecun1989optimal} & 94.17& 93.29 & 160 (T)\\
Jackpot (Ours)  & 94.03$\pm$0.12 & 92.54$\pm$0.11  & \textbf{10 (S)}      \\
\textbf{Jackpot (Ours)}  & \textbf{94.30$\pm$0.07} & \textbf{93.51$\pm$0.07} & 30 (S)     \\
\bottomrule
\end{tabular}}

\end{table}

\begin{table}[!t]
\setlength{\tabcolsep}{0.5em}
\setlength{\abovecaptionskip}{2pt}
\caption{\label{cifar100} Comparison with existing weight training methods for pruning VGGNet-19~\cite{simonyan2015very} and ResNet-32~\cite{he2016deep} on CIFAR-100~\cite{krizhevsky2009learning}. We report top-1 accuracy (\%), training cost (T) and searching cost (S).}
\centering
\resizebox{\columnwidth}{!}{
\begin{tabular}{c|cc|c}
\toprule
Pruning Rate& \quad\quad90\%\quad\quad  & \quad\quad95\%\quad\quad\quad\quad  & Epoch\\
\midrule
VGGNet-19~\cite{simonyan2015very} &  74.16& - & -  \\
SET~\cite{mocanu2018scalable} & 72.36 & 69.81 & 160 (T) \\
Deep-R~\cite{bellec2017deep} & 66.83 & 63.46   & 160 (T) \\
SNIP~\cite{lee2018snip} &  72.83 & 71.83 & 160 (T) \\
GraSP~\cite{wang2020picking} & 71.07 & 70.1   & 160 (T) \\
LT~\cite{frankle2018lottery} &  72.58 & 70.47  & 160 (T) \\
OBD~\cite{lecun1989optimal} & 73.83& 71.98 & 160 (T)\\
Jackpot (Ours)  & 72.41$\pm$0.14 & 72.27$\pm$0.21 & \textbf{10 (S)}  \\
\textbf{Jackpot (Ours)}  & \textbf{74.63$\pm$0.18} & \textbf{72.92$\pm$0.31} & 30 (S)    \\
\midrule                                
ResNet-32~\cite{he2016deep}    &      74.64    &  -      &       -       \\
SET~\cite{mocanu2018scalable} &  69.66 & 67.41 &  160 (T) \\
Deep-R~\cite{bellec2017deep} & 66.78 & 63.90  & 160 (T) \\
SNIP~\cite{lee2018snip} & 69.97 & 64.81&  160 (T) \\
GraSP~\cite{wang2020picking} & 70.12 & 67.05   &  160 (T) \\ 
LT~\cite{frankle2018lottery} & 69.63 & 66.48   & 160 (T) \\
OBD~\cite{lecun1989optimal} & 71.96& 68.73 & 160 (T)\\
Jackpot (Ours)  & 71.48$\pm$0.16 & 68.33$\pm$0.22  & \textbf{10 (S)}     \\
\textbf{Jackpot (Ours)}  & \textbf{72.68$\pm$0.13} & \textbf{70.01$\pm$0.14}  & 30 (S)    \\
  
\bottomrule
\end{tabular}}
\end{table}

\textbf{Implementation Details}.
We adopt the proposed SR-popup to searching for lottery jackpots using the SGD optimizer with an initial learning rate of 0.1 for all experiments.
With different total epochs for searching, \emph{i.e.}, 10 and 30, we adjust the learning rate with the cosine scheduler~\cite{loshchilov2016sgdr}.
The momentum is set to 0.9 and the batch size is set to 256.
The weight decay is set to 5 $\times 10^{-4}$ on CIFAR-10 and 1 $\times 10^{-4}$ on ImageNet.
For fair comparison, we conduct data augmentation for image pre-processing including cropping and horizontal flipping, which is the same as the official implementation in Pytorch~\cite{pytorch2015}. 
We repeat all of our experiments three times on CIFAR-10 with different seeds, and report the mean and standard deviation of top-1 classification accuracy.
On ImageNet, we run all experiments one time considering the heavy resource consumption and stable performance on the large dataset, and report both top-1 and top-5 classification accuracy.
All experiments are run on NVIDIA Tesla V100 GPUs.

\subsection{Comparison with Weight Training Methods}\label{experimentcomparison}
\textbf{CIFAR-10/100.}
On CIFAR-10/100, we compare lottery jackpots with many weight training competitors including OBD~\cite{lecun1989optimal}, SET~\cite{mocanu2018scalable}, Deep-R~\cite{bellec2017deep}, Lottery Tickets (LT)~\cite{frankle2018lottery}, SNIP~\cite{lee2018snip}, and GraSP~\cite{wang2020picking}. 
 We quantitatively report top-1 classification accuracy for pruning VGGNet-19 and ResNet-32 under two sparsity levels $\{90\%, 95\%\}$ in Table\,\ref{cifar10} and Table\,\ref{cifar100}.
The weight training or searching cost, which refers to epochs expenditure for obtaining the final pruned model, is also listed for comparison of the pruning efficiency.
Besides, we plot the performance of different approaches using the same training/search epoch at 95\% sparsity in Fig.\,\ref{fig:samecost}.
As can be observed in Table\,\ref{cifar10}, lottery jackpots provide a state-of-the-art performance while leveraging rare pruning computation cost on CIFAR-10.
For example, only 10 epochs are required for finding a lottery jackpot that contains only 10\% parameters of VGGNet-19, while attaining a superior top-1 accuracy of 93.70\%. 
In contrast, SNIP takes 160 epochs in total to reach a lower accuracy of 93.65\%.
Lottery jackpot also serves as a front-runner for pruning ResNet-32.
Different from the lottery tickets~\cite{frankle2018lottery} that need an entire weight training process for recovering performance, we directly search for lottery jackpots without modifying the pre-trained weights, resulting in even better pruned models (94.30\% \emph{v.s.} 92.61\% for top-1 accuracy) with more than 5$\times$ reductions on the pruning cost (30 epochs \emph{v.s.} 160 epochs). 

\begin{figure*}[!t]
\centering
\begin{subfigure}[t]{0.33\textwidth}
        \centering
        \includegraphics[width=\textwidth]{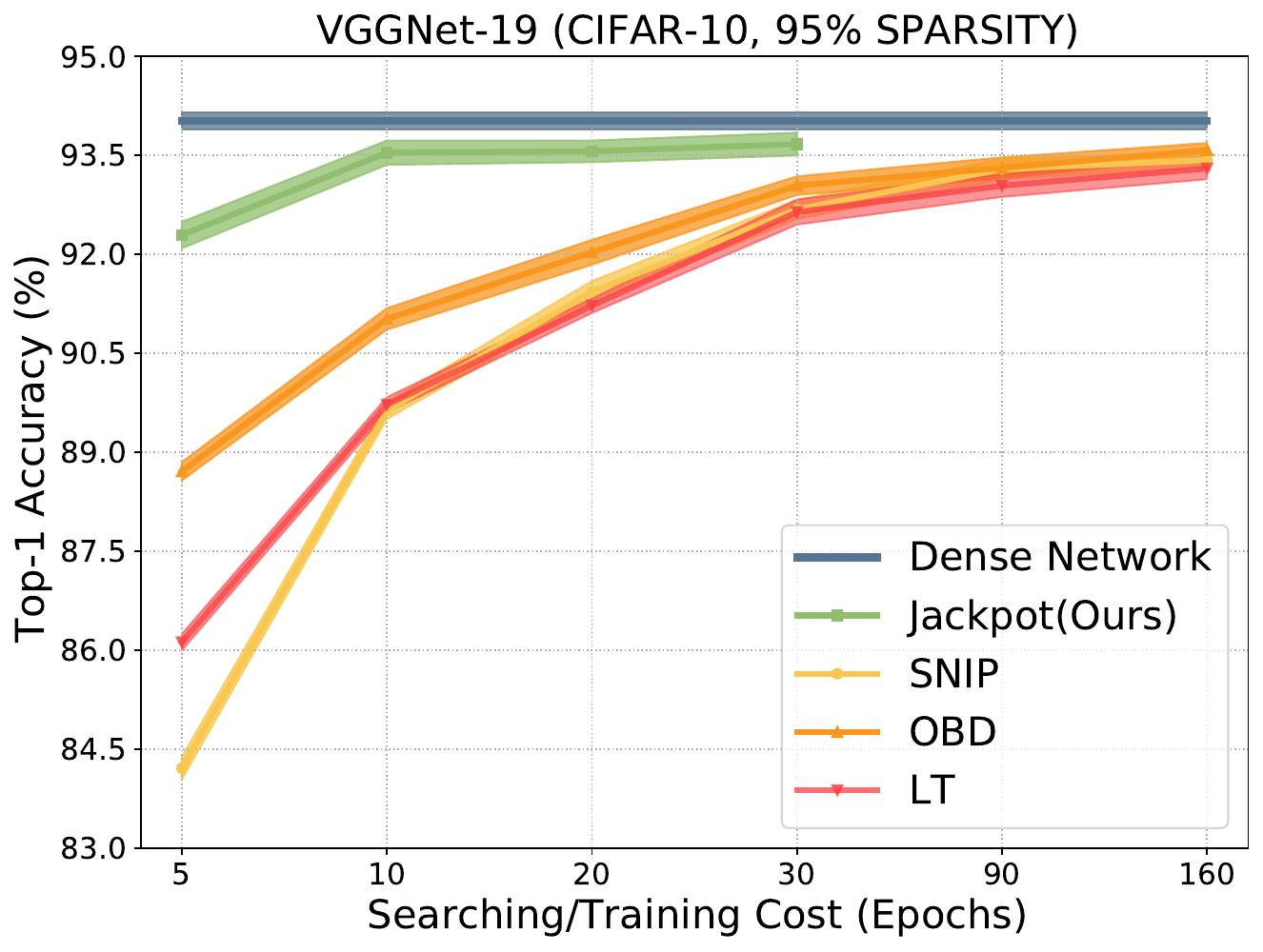}\\
\end{subfigure}
\begin{subfigure}[t]{0.33\textwidth}
        \centering
        \includegraphics[width=\textwidth]{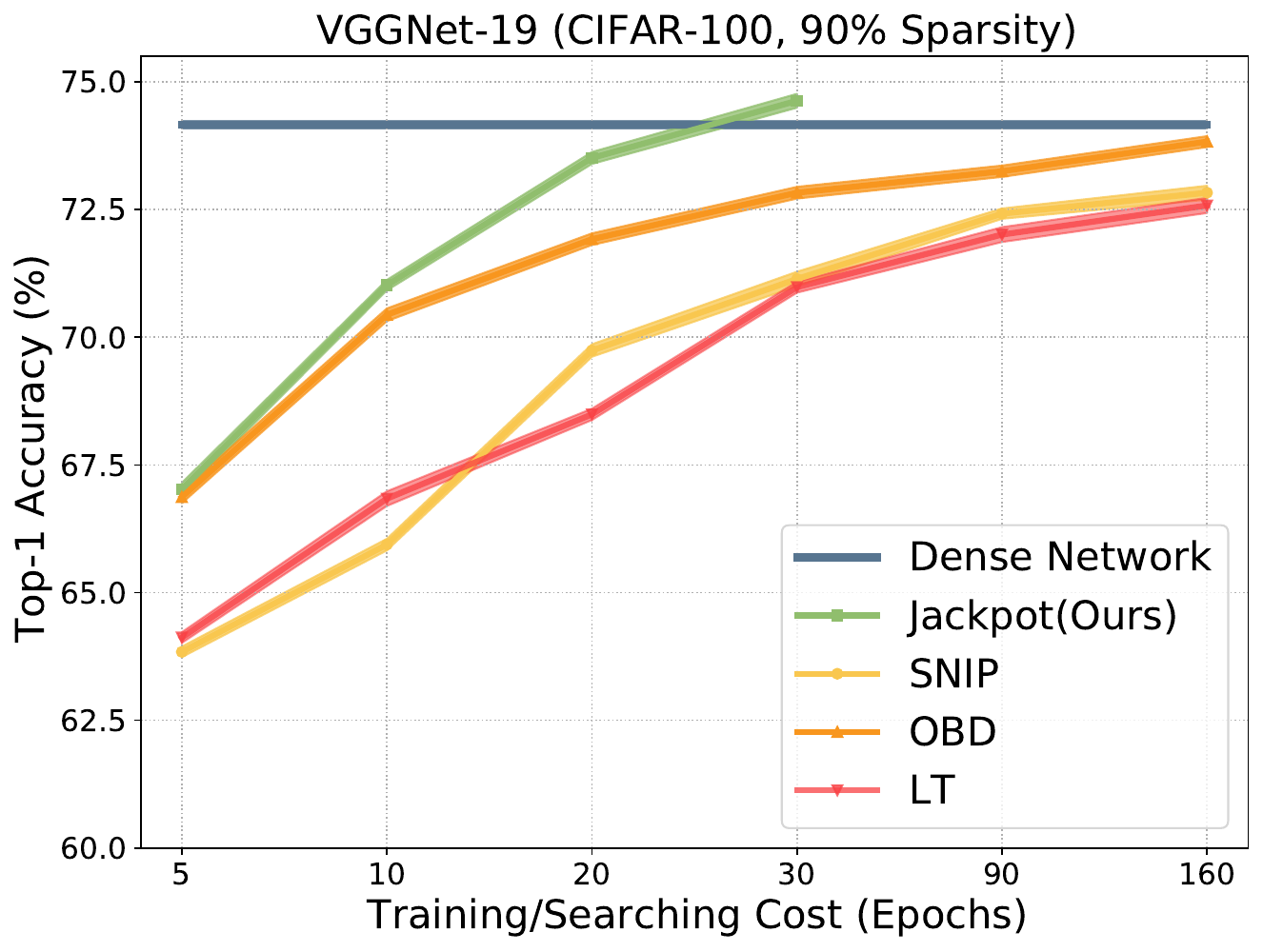}\\
\end{subfigure}
\begin{subfigure}[t]{0.33\textwidth}
        \centering
        \includegraphics[width=\textwidth]{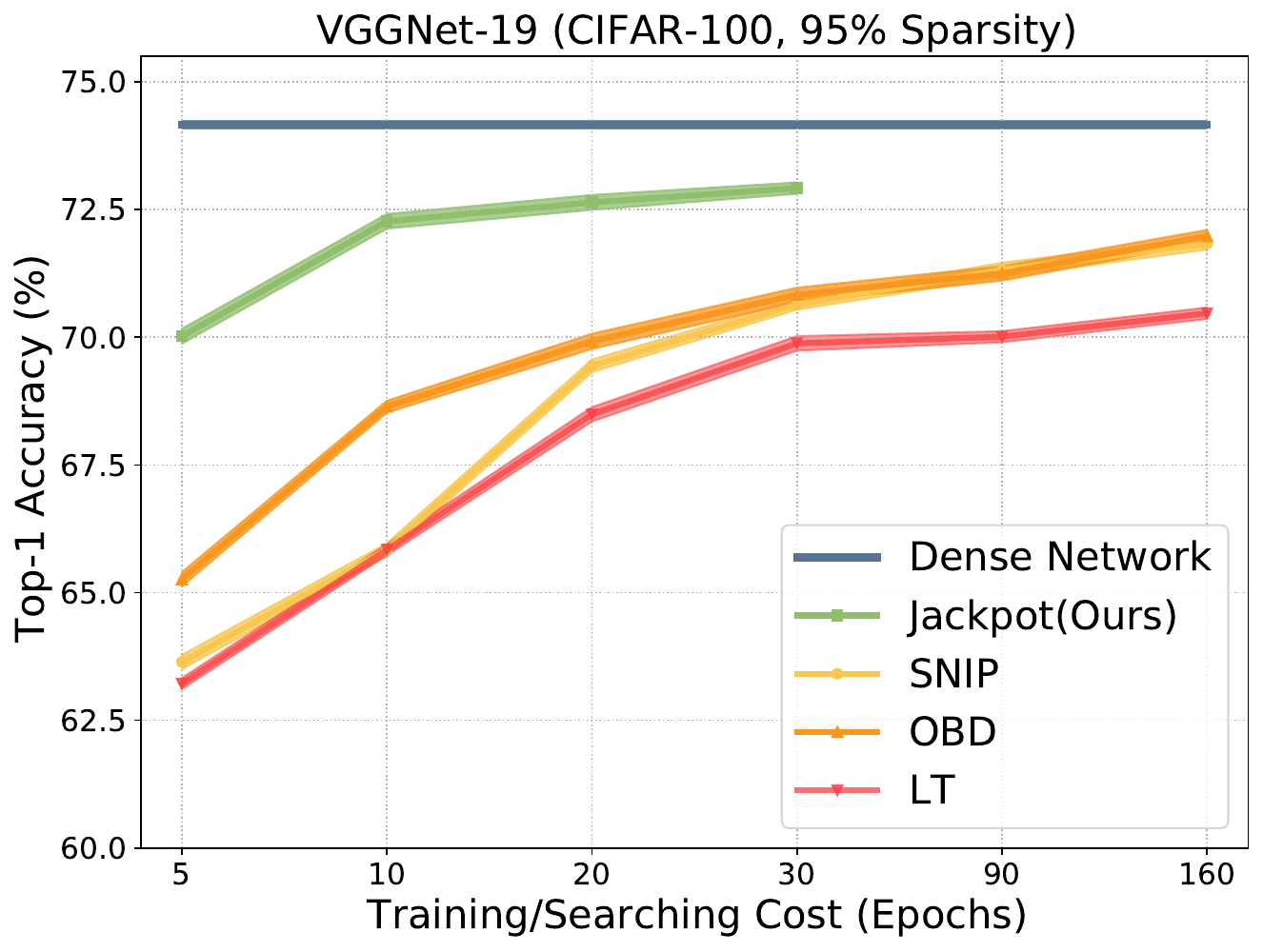}\\
\end{subfigure}
\begin{subfigure}[t]{0.33\textwidth}
        \centering
        \includegraphics[width=\textwidth]{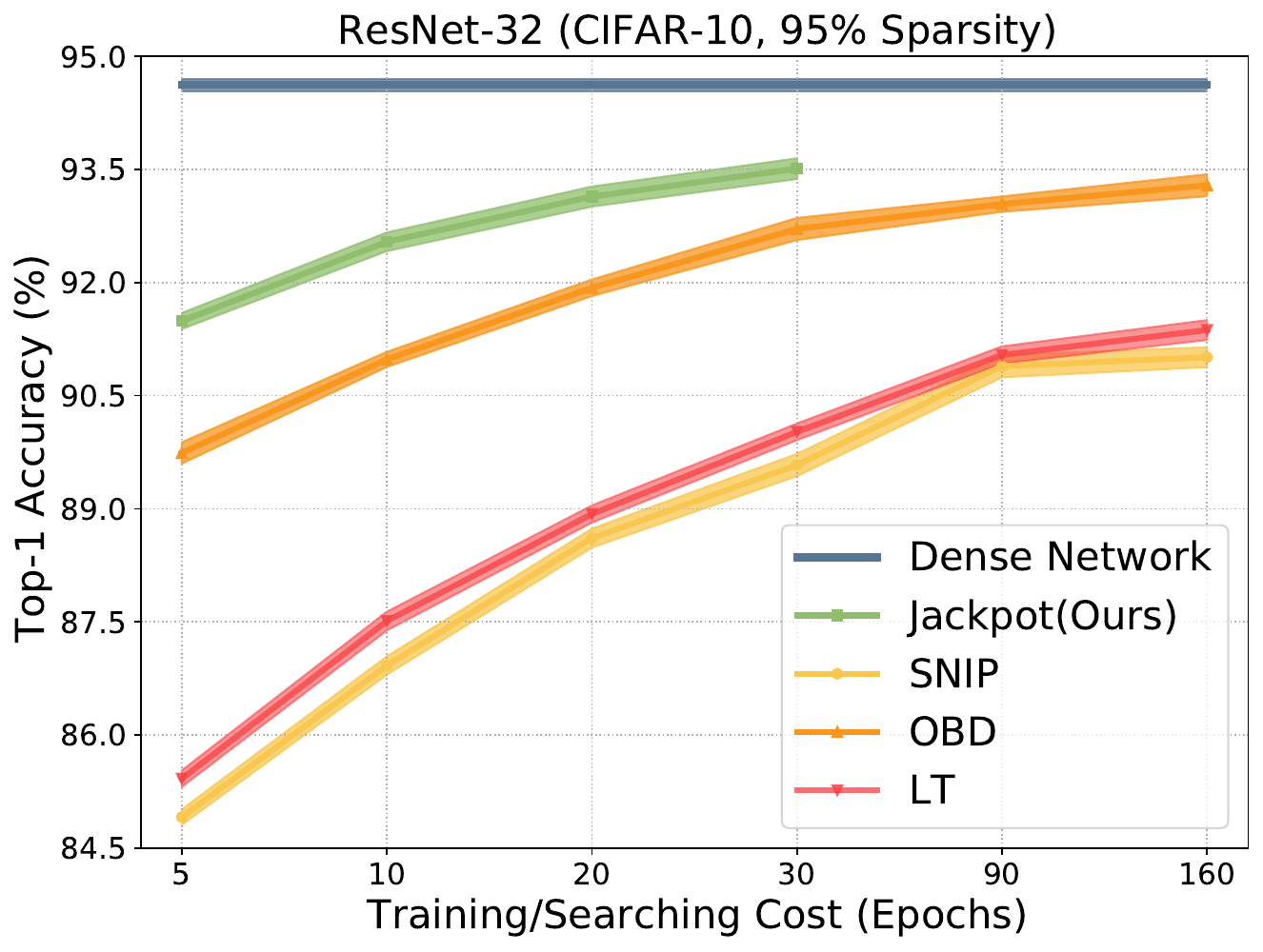}\\
 \end{subfigure}
\begin{subfigure}[t]{0.33\textwidth}
        \centering
        \includegraphics[width=\textwidth]{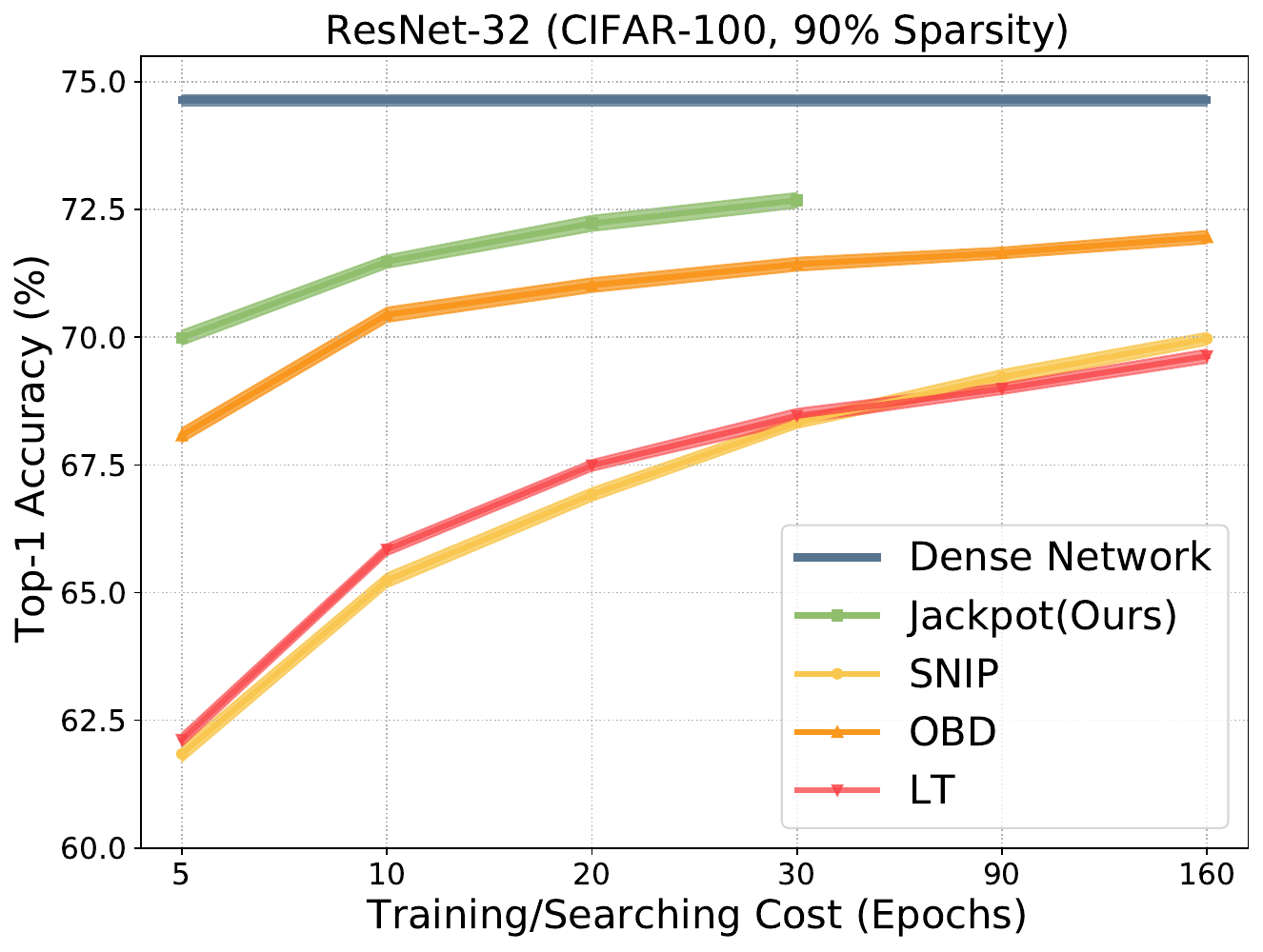}\\
\end{subfigure}
\begin{subfigure}[t]{0.33\textwidth}
        \centering
        \includegraphics[width=\textwidth]{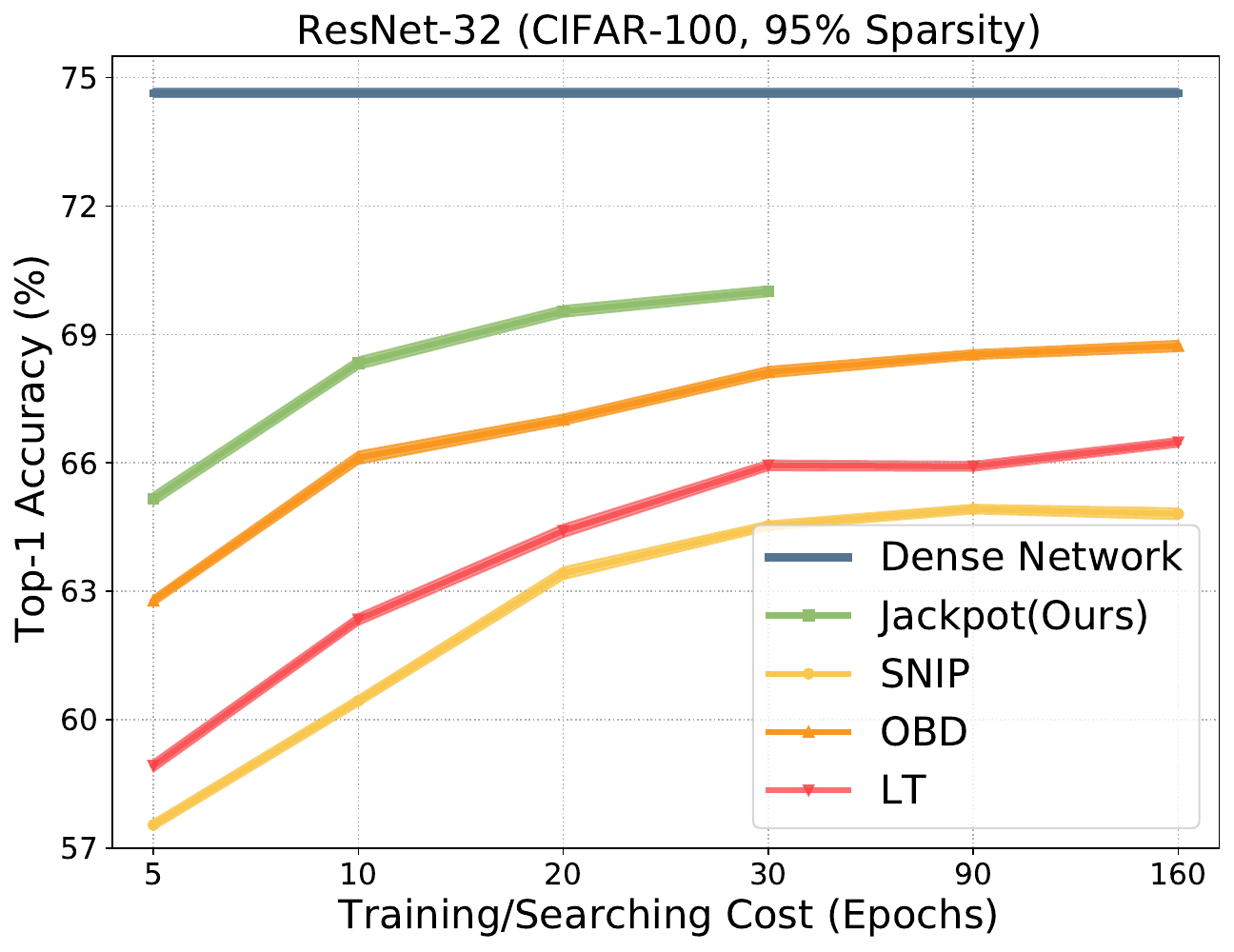}\\
\end{subfigure}
\centering
\caption{
Training/Search cost \emph{v.s.} top-1 accuracy of different methods including SNIP~\cite{lee2018snip}, OBD~\cite{lecun1989optimal} and LT~\cite{frankle2018lottery} to prune VGGNet-19~\cite{simonyan2015very} and ResNet-32~\cite{he2016deep} on CIFAR-10/100.}
\label{fig:samecost}
\end{figure*}

The results of CFIAR-100 in Table\,\ref{cifar100} also show that our lottery jackpots retain better top-1 accuracy than the competitors under different pruning rates with less cost. 
For instance, our method can achieve a top-1 accuracy of 74.63\% when pruning VGGNet-19 at 90\% sparsity, which surpasses other state-of-the-arts by a large margin of 2.27\%, 2.05\%, and 1.80\% higher than SET~\cite{mocanu2018scalable}, LT~\cite{frankle2018lottery}, and SNIP~\cite{lee2018snip}, respectively.
Note that it even exceeds the origin dense network by 0.47\%.
Compared with OBD~\cite{lecun1989optimal} that conducts fine-training after pruning the pre-trained ResNet-32, lottery jackpots can achieve better accuracy under all sparsity levels. 

\begin{table}[!t]
\centering
\setlength{\tabcolsep}{0.5em}
\setlength{\abovecaptionskip}{2pt}
\caption{\label{imagenet}Comparison with off-the-shelf weight training methods for pruning ResNet-50~\cite{he2016deep} on ImageNet~\cite{deng2009imagenet}. We report top-1 and top-5 accuracy (\%), training cost (T) and searching cost (S).}
\resizebox{\columnwidth}{!}{
\begin{tabular}{c|cc|c}
\toprule
   & Top-1 acc. & Top-5 acc. & Epoch    \\
\hline
Pruning Rate       & \multicolumn{2}{c|}{80\%}     &     \\
\hline
ResNet-50~\cite{he2016deep} & 76.15& 92.95 & -\\
SET~\cite{mocanu2018scalable} & 72.60 & 91.20 &100 (T) \\
Deep-R~\cite{bellec2017deep}  & 71.70 & 90.60  &100 (T) \\
Dynamic Sparse~\cite{mostafa2019parameter}  & 73.30 & 92.40   &100 (T) \\
SNIP~\cite{lee2018snip}  & 69.67 & 89.24   &100 (T) \\
GraSP~\cite{wang2020picking} & 72.06 & 90.82 &  150 (T) \\
RigL~\cite{evci2020rigging}  & 74.60 & -  &100 (T) \\
OBD~\cite{lecun1989optimal} & 75.12& 68.73 & 100(T)\\
Jackpot (Ours) &75.19 & 92.52 & \textbf{30 (S)}\\
\textbf{Jackpot (Ours)} & \textbf{75.71} & \textbf{92.90} & 60 (S)\\
\hline
Pruning Rate    & \multicolumn{2}{c|}{90\%}     &     \\
\hline
ResNet-50~\cite{he2016deep} & 76.15& 92.95 & -\\
SET~\cite{mocanu2018scalable}  & 70.40&90.10   &100( T) \\
Deep-R~\cite{bellec2017deep}  & 70.20 & 90.00 &100 (T) \\
Dynamic Sparse~\cite{mostafa2019parameter}    & 71.60 & 90.50 &100 (T) \\
SNIP~\cite{lee2018snip}   & 67.21 & 87.19   &100 (T) \\
GraSP~\cite{wang2020picking}  & 68.14 &88.67& 150 (T) \\
RigL~\cite{evci2020rigging}  & 72.00  & - &100 (T) \\
OBD~\cite{lecun1989optimal} & 72.51& 90.78 & 100(T)\\
Jackpot (Ours) & 72.61 & 91.09 & \textbf{30 (S)}\\
\textbf{Jackpot (Ours)} & \textbf{73.04} & \textbf{91.34} & 60 (S)\\
\bottomrule
\end{tabular}}
\end{table}

\begin{table}[!t]

\setlength{\tabcolsep}{0.5em}
\setlength{\abovecaptionskip}{2pt}
\caption{Comparison with existing weight training methods for pruning MobileNet-V1~\cite{howard2017mobilenets} and EfficientNet-B4~\cite{tan2019efficientnet} on ImageNet~\cite{deng2009imagenet}. We report top-1 accuracy (\%), training cost (T) and searching cost (S).}
\centering
\resizebox{\columnwidth}{!}{
\begin{tabular}{c|cc|c}
\toprule
&  Top-1 acc.\quad & \quad Top-5 acc. & Epoch    \\
\hline
\ Pruning Rate       & \multicolumn{2}{c|}{80\%}     &     \\
\hline
\;MobileNet-V1~\cite{howard2017mobilenets}\; &  71.93& 90.37 & -  \\
SNIP~\cite{lee2018snip} & 63.95 &85.12 &180(T) \\
GraSP~\cite{wang2020picking} & 63.71 & 84.99  & 180(T) \\
RigL~\cite{evci2020rigging} &  65.01 &  86.98  &180 (T) \\
Jackpot (Ours)  & 67.21 & 87.83 & \textbf{30 (S)}     \\
\textbf{Jackpot (Ours)}  & \textbf{67.50} & \textbf{88.02}  & 60 (S)    \\
\midrule                                
EfficientNet-B4~\cite{tan2019efficientnet}   &      82.61    &  96.41      &       -     \\
SNIP~\cite{lee2018snip} & 74.81 & 92.12 & 180(T) \\
GraSP~\cite{wang2020picking} & 74.70 & 92.01 &180(T) \\
RigL~\cite{evci2020rigging} & 75.46& 92.57  & 180 (T) \\
Jackpot (Ours)  & 76.26 & 93.14 & \textbf{30 (S)}     \\
\textbf{Jackpot (Ours)}  & \textbf{76.67} & \textbf{93.39}  & 60 (S)    \\
\bottomrule
\end{tabular}}
\label{lightweight}
\end{table}

From Fig.\,\ref{fig:samecost}, we further observe supreme performance of lottery jackpots when maintaining similar pruning cost.
Detailedly, previous methods~\cite{lee2018snip, wang2020picking, evci2020rigging} suffer significant performance degradation with limited epochs, mostly due to the insufficient training of weights.
On the contrary, we directly search lottery jackpots on the pre-trained weights, leading to a satisfying performance with far less cost.
%

\begin{table}[!t]

\centering
\setlength{\tabcolsep}{0.5em}
\setlength{\abovecaptionskip}{2pt}
\caption{Comparison with existing weight searching methods for pruning ResNet-50~\cite{he2016deep} on ImageNet~\cite{deng2009imagenet}. The pruning rate is set to 90\%. We report top-1 and top-5 accuracy (\%), and searching cost (S).}
\resizebox{\columnwidth}{!}{
\begin{tabular}{c|cc|c}
\toprule
   & Top-1 acc. & Top-5 acc. & Epoch    \\
\midrule
ResNet-50~\cite{he2016deep} & 76.15& 92.95 &- \\
Zhou~\emph{et al.}~\cite{mocanu2018scalable}  & 65.41&86.11   & \textbf{30(S)}\\
Edge-popup~\cite{bellec2017deep}  & 70.20 & 88.91 & \textbf{30(S)} \\
\textbf{SR-popup (Ours)} & \textbf{72.61} & \textbf{91.09} & \textbf{30(S)}\\
\midrule
ResNet-50~\cite{he2016deep} & 76.15& 92.95 &- \\
Zhou~\emph{et al.}~\cite{mocanu2018scalable}  & 68.40 &88.03   & \textbf{60(S)} \\
Edge-popup~\cite{bellec2017deep}  & 70.89 & 90.12 & \textbf{60(S)} \\
\textbf{SR-popup (Ours)} & \textbf{73.04} & \textbf{91.34} &\textbf{60(S)}\\
\bottomrule
\end{tabular}}
\label{weight_search}
\end{table}

\textbf{ImageNet}.
We further demonstrate the efficacy of our lottery jackpots for pruning ResNet-50 on the challenging ImageNet dataset.
Two pruning rates $\{80\%, 90\%\}$ are considered for fair comparison with other state-of-the-arts including SET~\cite{mocanu2018scalable}, Deep-R~\cite{bellec2017deep}, Dynamic Sparse~\cite{mostafa2019parameter}, SNIP~\cite{lee2018snip}, GraSP~\cite{wang2020picking}, OBD~\cite{lecun1989optimal} and RigL~\cite{evci2020rigging}.
In Table\,\ref{imagenet}, lottery jackpots beat all the competitors in both top-1 and top-5 accuracy under the same sparsity ratio.
We can search for a lottery jackpot with only 1/5 computation cost compared with GraSP~\cite{wang2020picking} (30 epochs for us and 150 epochs for GraSP), while obtaining pruned model of 80\% pruning rate with higher top-1 and top-5 accuracy (75.19\% and 92.62\% for lottery jackpot \emph{v.s.} 72.06\% and 90.82\% for GraSP).
When increasing the sparsity level to 90\%, a lottery jackpot that achieves 73.04\% in top-1 accuracy with 5.80\% and 1.04\% improvements over GraSP~\cite{wang2020picking} and RigL~\cite{evci2020rigging}, can be successfully searched with fewer epochs (60, 100, and 150 epochs for lottery jackpots, RigL and GraSP, respectively). These results demonstrate the feasibility of directly finding good-performing subnets without the requirement of time-consuming weight training process.

We also conduct experiments for pruning light-designed networks, including MobileNet-V1~\cite{howard2017mobilenets} and EfficientNet-B4~\cite{tan2019efficientnet} at a pruning rate of 80\%. 
Table\,\ref{lightweight} lists the performance of lottery jackpots in comparison with SNIP~\cite{lee2018snip}, GraSP~\cite{wang2020picking}, and RigL~\cite{evci2020rigging} based on our re-implementation.
The results again suggest the superiority of lottery jackpots for the higher top-1 accuracy even without modifying any pre-trained weights.
Upon MobileNet-V1, a lottery jackpot that significantly outperforms the RigL by 2.21\% can be found with 6$\times$ reduction on the pruning cost.
For the more compact EfficientNet-B4, we observe a considerable performance drop for all methods.
Nevertheless, lottery jackpots still performs the best. 
Thus, it well demonstrates the advantage of finding lottery jackpots in pre-trained light-weight networks empowered with the ability to effectively reduce the network complexity.

\subsection{Comparison with Weight Searching Methods}\label{experimentsearch}
In this section, we compare the performance of lottery jackpots found by our proposed SR-popup and weight-searching methods of Zhou~\emph{et al.}~\cite{zhou2019deconstructing} and Edge-popup~\cite{ramanujan2020s} that are originally designed for searching good-performing subnets in randomly initialized networks.
Results for pruning pre-trained ResNet-50 on ImageNet are listed in Table\,\ref{weight_search}.
Our proposed SR-popup significantly outperforms other methods when consuming the same searching epochs.
Particularly, serious performance drops are observed for all methods except our SR-popup with a limited epoch of 30, which demonstrates the efficacy of leveraging magnitude-based pruning as a warm-up for the relaxed masks.
Although adding the searching epochs brings clear improvement for Edge-popup, its performance still falls behind SR-popup by a noticeable margin.
For example, when consuming 60 searching epochs, SR-popup successfully locates a lottery jackpot with 73.04\% top-1 accuracy, surpassing edge-popup by 2.15\% whose searching process is heavily stumbled by the distortion error of weight interdependence as discussed in Sec.\,\ref{interdependence}.
%

\begin{figure}[!t]
\begin{center}
\includegraphics[height=0.62\linewidth]{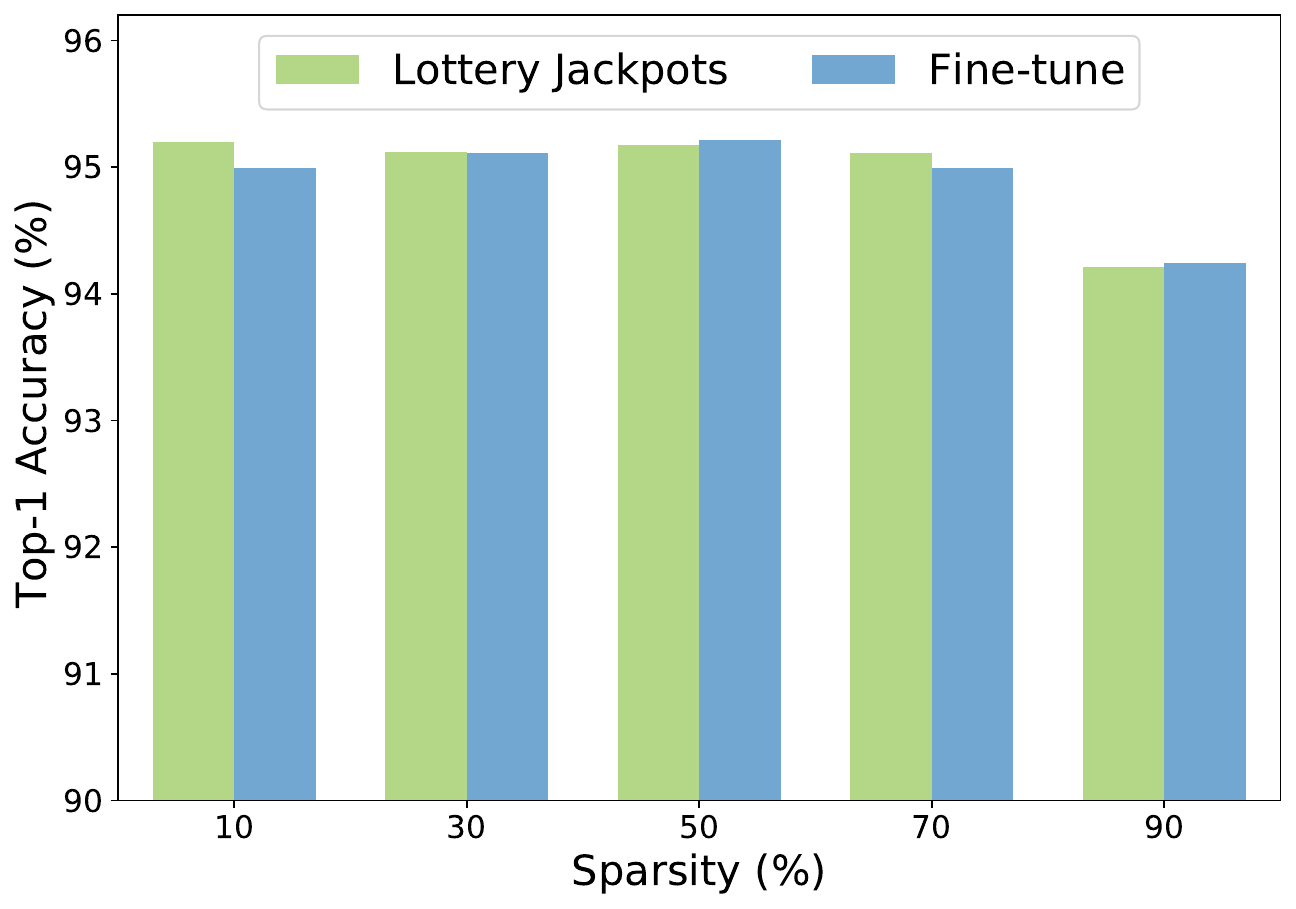}
\end{center}
\caption{\label{fig7}
Comparison between the searched lottery jackpots and fine-tuned results based on the searched mask under different sparsity levels for pruning ResNet-32~\cite{he2016deep} on CIFAR-10~\cite{krizhevsky2009learning}.
}
\end{figure}
\subsection{Performance Analysis}\label{ablation}

In this section, we conduct multiple experiments to investigate the performance of lottery jackpots. We first study the effect of weight training process for lottery jackpots. After searching the lottery jackpots, we freeze the found mask and fine-tune the weights to investigate whether such a weight training process can further benefit the performance of the pruned model or not. The weight training objective is formulated as:
\begin{equation}\label{eq9}
\begin{split}
    \min_{\tilde{\mathbf{w}}} \; \mathcal{L} (h(\mathbf{\hat{m}}) \odot \tilde{\mathbf{w}} \; ; \mathcal{D}).
\end{split}
\end{equation}

We conduct experiments for ResNet-32 on CIFAR-10 with different pruning rates. 
The weights are fine-tuned with 150 epochs and the other settings are described in Sec.\,\ref{setting}. 
Fig.\,\ref{fig7} shows that the improvement of weight training process is negligible. 
It can not further improve the performance of lottery jackpots, and even lead to a little accuracy degradation under high sparsity.
Thus, we can conclude that the time-consuming weight training process is unnecessary for getting good-performing sparse models to some extent.
In contrast, lottery jackpots already exist in pre-trained models.

\begin{figure}[!t]
\begin{center}
\includegraphics[height=0.7\linewidth]{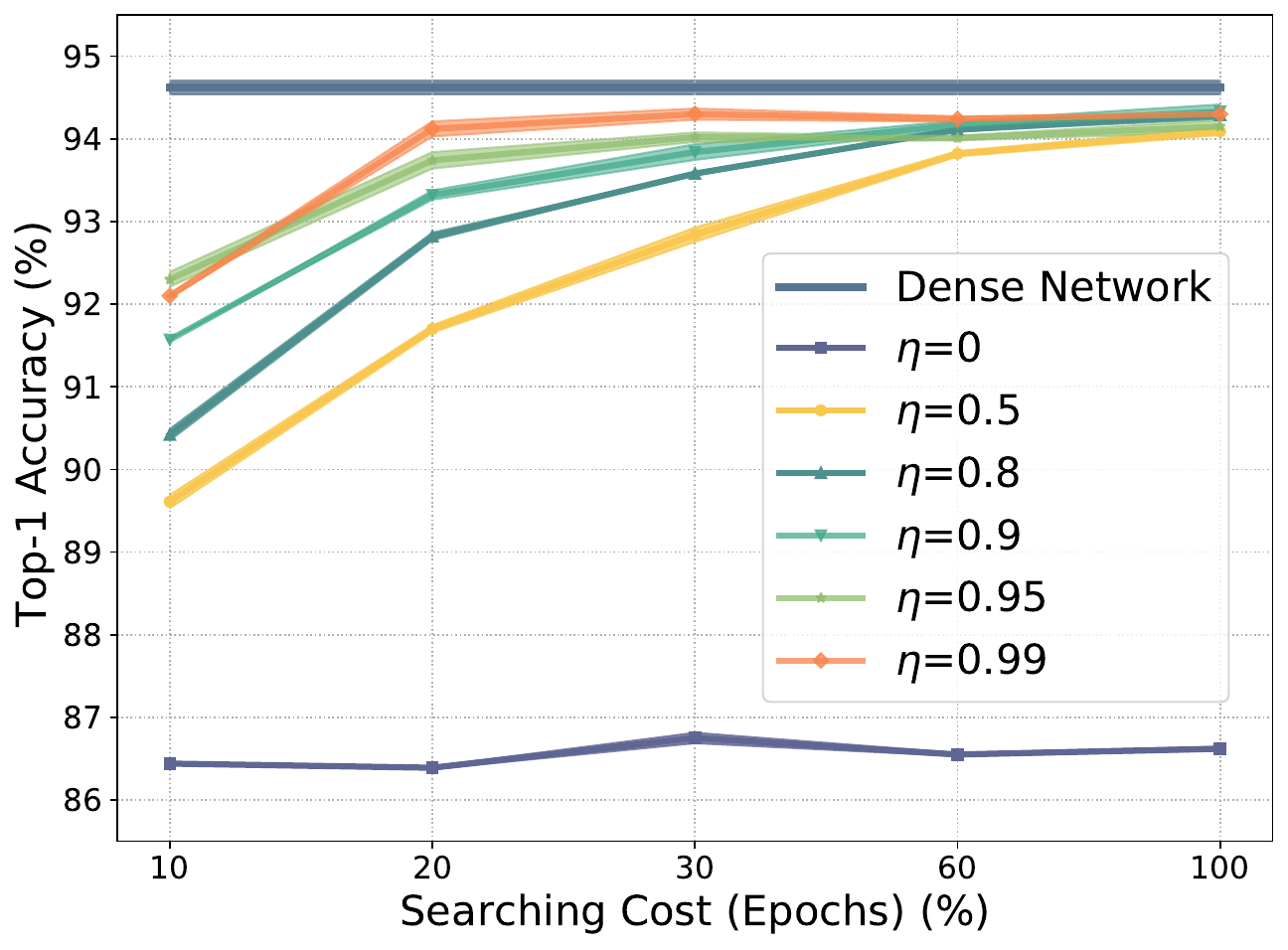}
\end{center}
\caption{\label{ablation:0}
Ablation study for the initialization of relaxed masks corresponding to pruned weights.
}
\end{figure}

Then, we investigate the influence of searching epochs for lottery jackpots.
Detailedly, while comparing the performance, we also observe the sparse architecture overlap between lottery jackpots and the initial mask $\hat{\mathbf{m}}$ obtained by magnitude pruning.
We show the quantitative results in Table\,\ref{searchepoch}.
As the searching epoch begins to increase, the performance of found lottery jackpots can be boosted and the sparse architecture also changes to some extent.
However, such a phenomenon is not consistent with more searching epochs.
The performance of lottery jackpots searched with 100 and 1000 epochs are at the same level.
Further, the sparse structure of lottery jackpots gradually tends to be stable, and will not change significantly with the increase of searching cost.
Thus, only a relatively small amount of computation cost is needed to find lottery jackpots, which well demonstrates the efficiency of our approach.

\begin{table}[!t]
\setlength{\tabcolsep}{0.7em}
\setlength{\abovecaptionskip}{2pt}
\caption{\label{searchepoch}Top-1 accuracy (\%) of lottery jackpots with different epochs and sparse architecture overlap with the magnitude-based pruning for pruning ResNet-32~\cite{he2016deep} on CIFAR-10~\cite{krizhevsky2009learning} at 90\% sparse level.}
\centering
\resizebox{0.85\columnwidth}{!}{
\begin{tabular}{ccc}
\toprule
Search epoch & Top-1  acc. & Architecture\\
 &   & overlap\\
\midrule
\textbf{0} & 10.00$\pm$0.01 & 100.0\% \\
10 & 94.03$\pm$0.01& 96.8\% \\
30 & 94.30$\pm$0.08 & 96.1\% \\
60 & \textbf{94.43$\pm$0.13} & 95.4\% \\
100 & 94.41$\pm$0.11 & 94.2\% \\
300 & 94.42$\pm$0.11 & 94.3\% \\
1000 & 94.43$\pm$0.14 & 94.1\% \\
\bottomrule
\end{tabular}}
\end{table}


Lastly, we perform four ablation studies for the proposed SR-popup method. The experiments are conducted for pruning ResNet-32 and VGGNet-19 on CIFAR-10 at 90\% sparse level using 30 searching epochs.
1) Table\,\ref{MSablation} shows the performance comparison between without using magnitude-based initialization (w/o W) or without using short-restriction (w/o S), and using both of them (SR) to search lottery jackpots.
The results suggest that both components in the proposed SR-popup are essential to the performance of lottery jackpots.
2) For the magnitude-based initialization in Eq.\,(\ref{ini_func}), we investigate how the initial value $\eta$ of the relaxed masks affects the pruned weights. 
Detailedly, we plot the top-1 accuracy~\emph{v.s.} searching epochs using different $\eta$ to search lottery jackpots.
Fig.\,\ref{ablation:0} shows that smaller initial values result in a slower decrease in the training loss. To explain, the distance between the masks corresponding to the pruned and preserved weights is too large, resulting in slow convergence. 
%
%
3) We excavate the choice for the elected weight swapping pairs. 
Two variants include selecting these with the smallest loss drops random selection are considered for comparison with our proposed method. 
Results in Table\,\ref{ablation:1} suggest that our proposed restriction method surpasses other variants by a large margin, which demonstrates our point for only preserving weight swapping that contributes the most significant expected loss drops for searching lottery jackpots.
4) The declining schedule for the restriction number of weight swapping,~\emph{i.e.}, Eq.\,(\ref{eq:schedule}), is 
compared with its inverse version:
\begin{equation}\label{eq:schedule2}
   q_t = \lceil \left\|\Psi^*_t\right\|_0 (\frac{t}{t_f})^4\rceil,
\end{equation}
and a constant schedule is also considered,~\emph{w.r.t.} $q_t = 1$. 
This circumstance is equivalent to adopting the edge-popup algorithm on the foundation of our proposed magnitude-based initialization. 
The results listed in Table\,\ref{ablation:2} suggest that our proposed schedule leads to the best performance thanks to a sufficient exploration for weight swapping during the early searching iterations and incremental restriction of unstable weight swapping for efficient convergence.

\begin{table}[!t]
\setlength{\tabcolsep}{0.7em}
\setlength{\abovecaptionskip}{2pt}
\caption{\label{MSablation}Ablation study for the components in our proposed SR-popup.}
\centering
\resizebox{\columnwidth}{!}{
\begin{tabular}{c|ccc}
\toprule
Method & w/o M & w/o S & SR \\
\midrule
Top-1 acc. & 94.01$\pm$0.11 & 93.69$\pm$0.07 & \textbf{94.30$\pm$0.07}  \\
\bottomrule
\end{tabular}}
\end{table}

\begin{table}[!t]
\setlength{\tabcolsep}{0.7em}
\setlength{\abovecaptionskip}{2pt}
\caption{\label{ablation:1}Ablation study for the choice of weight swapping pairs.}
\centering
\resizebox{\columnwidth}{!}{
\begin{tabular}{c|ccc}
\toprule
Method & Inverse & Random & Ours \\
\midrule
Top-1 acc. & 93.13$\pm$0.12 & 93.84$\pm$0.08 & \textbf{94.30$\pm$0.07}  \\
\bottomrule
\end{tabular}}
\end{table}

\begin{table}[!t]

\setlength{\tabcolsep}{0.7em}
\setlength{\abovecaptionskip}{2pt}
\caption{\label{ablation:2}Ablation study for the schedule of restriction number for weight swapping.}
\centering
\resizebox{\columnwidth}{!}{
\begin{tabular}{c|ccc}
\toprule
Schedule & Constant & Inverse & Ours \\
\midrule
Top-1 acc. & 93.13$\pm$0.12 & 93.44$\pm$0.08 & \textbf{93.51$\pm$0.14}  \\
\bottomrule
\end{tabular}}
\end{table}

\section{Limitations}
As stressed throughout the paper, our lottery jackpots are proposed on the premise of pre-trained models, which, we believe, are mostly available from the Internet or the client. However, the opportunity exists to be inaccessible to the pre-trained models in some situation, indicating the inapplicability of our lottery jackpots. Besides, our limited hardware resources disable us to explore the existence of lottery jackpots beyond the convolutional neural networks. We expect to show more results on other tasks such as natural language processing in our future work.

\section{Discussion and Conclusion}  
In the field of network pruning, the state of pre-trained models has been increasingly overlooked. 
Instead, many researchers devised complex and time-consuming mechanisms for training sparse networks from scratch.
On the contrary, in this paper, we re-justify the importance of pre-trained models by revealing the existence of lottery jackpots that, high-performing sub-networks emerge in pre-trained models without the necessity of weight training.
To mitigate the inefficiency problem of existing searching methods, we further present a novel SR-popup method that enhances both the initialization and searching process of lottery jackpots.
For the mask initialization, we experimentally observe that directly leveraging existing pruning criteria leads to sparse masks that overlap with our lottery jackpot to a notable extent.
Among those pruning criteria, the magnitude-based pruning results in the most similar masks with lottery jackpots.
Based on this insight, we initialize the sparse mask using magnitude pruning.
On the other hand, a short restriction mechanism is proposed to restrict change of masks that may have potential negative impacts on the training loss during searching under theoretical guarantee.
Extensive experiments demonstrate that our lottery jackpots can achieve comparable or even better performance with many state-of-the-arts without complex expert knowledge for training sparse networks, while greatly reducing the pruning cost.
%
%
%

%
%
%


\section*{Acknowledgement}
This work was supported by National Key R\&D Program of China (No.2022ZD0118202), the National Science Fund for Distinguished Young Scholars (No.62025603), the National Natural Science Foundation of China (No. U21B2037, No. U22B2051, No. 62176222, No. 62176223, No. 62176226, No. 62072386, No. 62072387, No. 62072389, No. 62002305 and No. 62272401), and the Natural Science Foundation of Fujian Province of China (No.2021J01002,  No.2022J06001).


%



\ifCLASSOPTIONcaptionsoff
  \newpage
\fi



%



\bibliographystyle{IEEEtran}
\bibliography{main}

\ifCLASSOPTIONcaptionsoff
  \newpage
\fi

%

%


\begin{IEEEbiography}[{\includegraphics[width=1in,height=1.25in,clip,keepaspectratio]{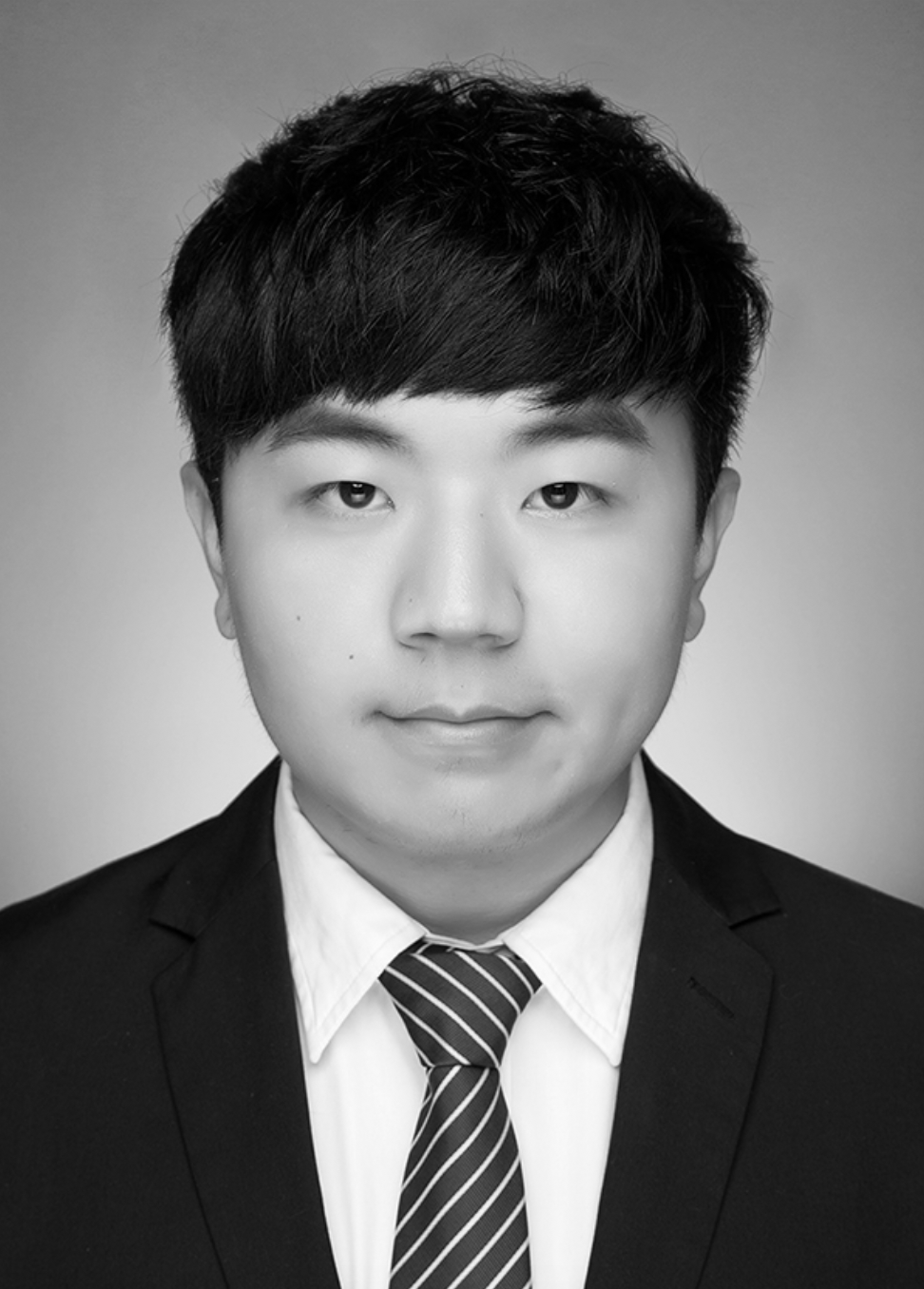}}]{Yuxin Zhang} received the B.E. degree in Computer Science, School of Informatics, Xiamen University, Xiamen, China, in 2020.
He is currently pursuing the P.H.D degree with Xiamen University, China. His publications on top-tier conferences/journals include IEEE TPAMI, IEEE TNNLS, NeurIPS, ICLR, ICML, ICCV, IJCAI and so on. His research interests include computer vision and neural network compression \& acceleration.
\end{IEEEbiography}

\begin{IEEEbiography}[{\includegraphics[width=1in,height=1.25in,clip,keepaspectratio]{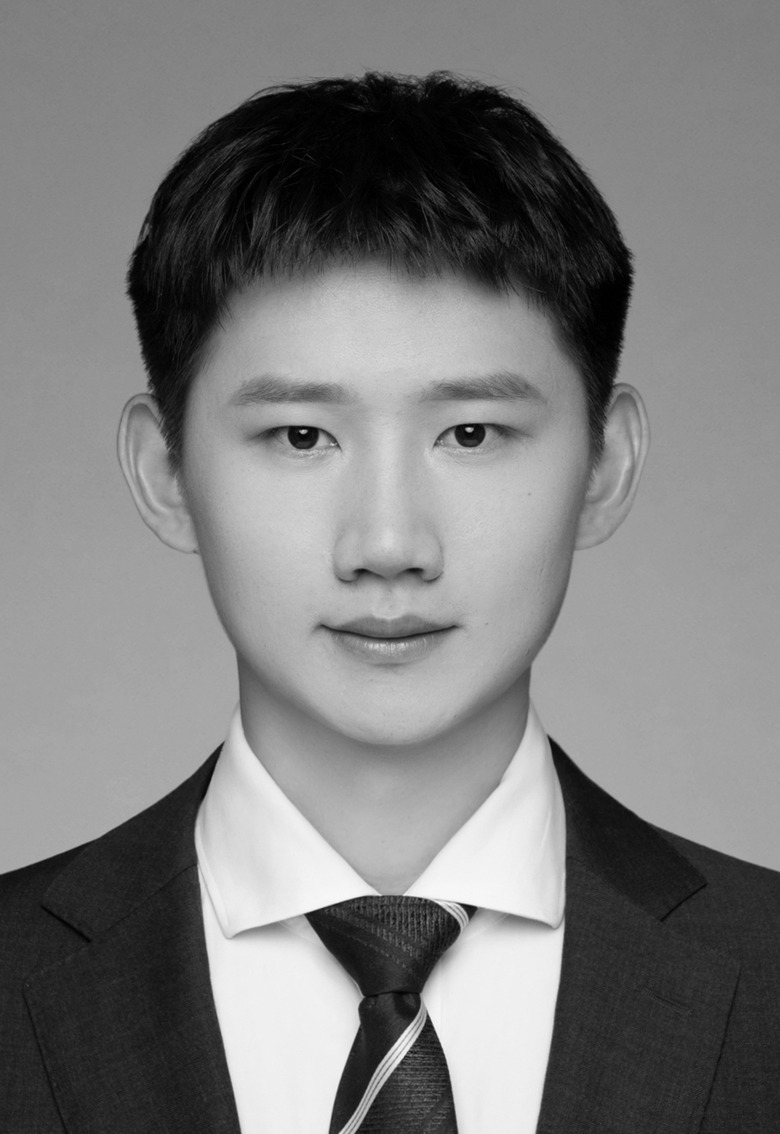}}]{Mingbao Lin} finished his M.S.-Ph.D. study and obtained the Ph.D. degree in intelligence science and technology from Xiamen University, Xiamen, China, in 2022. Earlier, he received the B.S. degree from Fuzhou University, Fuzhou, China, in 2016.

He is currently a senior researcher with the Tencent Youtu Lab, Shanghai, China. His publications on top-tier conferences/journals include IEEE TPAMI, IJCV, IEEE TIP, IEEE TNNLS, CVPR, NeurIPS, AAAI, IJCAI, ACM MM and so on. His current research interest is to develop efficient vision model, as well as information retrieval.
\end{IEEEbiography}

\begin{IEEEbiography}[{\includegraphics[width=1in,height=1.25in,clip,keepaspectratio]{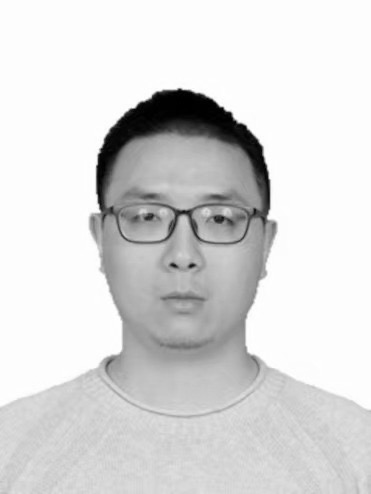}}]{Yunshan Zhong} received the B.Sc degree in Software Engineering from the Beijing Institute of Technology, Beijing, China in 2017, and the M.S. degree in Software Engineering from Peking University, Beijing, China in 2020. He is currently a second-year Ph.D. student in the MAC lab, the Institute of Artificial Intelligence, Xiamen University, China, under the supervision of Prof. Rongrong Ji. He has published multiple peer-reviewed papers on top-tier conferences including CVPR, ICCV, and ECCV. His current research interest is model compression.
\end{IEEEbiography}

\begin{IEEEbiography}[{\includegraphics[width=1in,height=1.25in,clip,keepaspectratio]{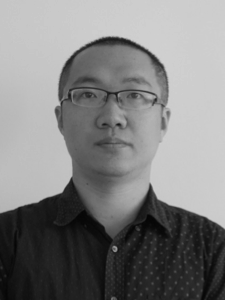}}]{Fei Chao} received the B.Sc. degree in mechanical engineering
from the Fuzhou University, P. R. China, and the M.Sc. Degree
with distinction in computer science from the University of
Wales, UK, in 2004 and 2005, respectively, and the Ph.D. degree
in robotics from the Aberystwyty University, Wales, UK in 2009.
He was a Research Associate under the supervision of Professor
Mark H. Lee at the Aberystwyth University from 2009 to 2010.
He is currently an Assistant Professor with the Cognitive Science
Department, at the Xiamen University, P. R. China.
He has published about 20 peer-reviewed journal and conference papers. His
research interests include developmental robotics, machine learning, and optimization algorithms. He is the Vice Chair of the IEEE Computer Intelligence Society Xiamen Chapter. Also, he is a member of ACM and CCF.
\end{IEEEbiography}

\begin{IEEEbiography}[{\includegraphics[width=1in,height=1.25in,clip,keepaspectratio]{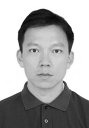}}]{Rongrong Ji}
(Senior Member, IEEE) is a Nanqiang Distinguished Professor at Xiamen University, the Deputy Director of the Office of Science and Technology at Xiamen University, and the Director of Media Analytics and Computing Lab. He was awarded as the National Science Foundation for Excellent Young Scholars (2014), the National Ten Thousand Plan for Young Top Talents (2017), and the National Science Foundation for Distinguished Young Scholars (2020). His research falls in the field of computer vision, multimedia analysis, and machine learning. He has published 50+ papers in ACM/IEEE Transactions, including TPAMI and IJCV, and 100+ full papers on top-tier conferences, such as CVPR and NeurIPS. His publications have got over 10K citations in Google Scholar. He was the recipient of the Best Paper Award of ACM Multimedia 2011. He has served as Area Chairs in top-tier conferences such as CVPR and ACM Multimedia. He is also an Advisory Member for Artificial Intelligence Construction in the Electronic Information Education Committee of the National Ministry of Education.
\end{IEEEbiography}




\end{document}